\documentclass{article}
\usepackage[OT1]{fontenc}
\usepackage{iclr2027_conference,times}

\usepackage{hyperref}
\usepackage{url}
\usepackage{amsmath}
\usepackage{amssymb}
\usepackage{graphicx}
\usepackage{booktabs}
\usepackage{multirow}
\usepackage{xcolor}
\usepackage{colortbl}
\usepackage{enumitem}
\usepackage{algorithm}
\usepackage{algorithmic}
\usepackage[capitalize,noabbrev]{cleveref}
\usepackage{subcaption}
\usepackage[percent]{overpic}%
\usepackage[export]{adjustbox}%
\definecolor{customcolor}{HTML}{004496}
\definecolor{custompink}{HTML}{FF69B4}

\hypersetup{
    colorlinks=true,
    linkcolor=red,
    citecolor=customcolor,
    urlcolor=custompink
}

\newcommand{\x}{\mathbf{x}}
\newcommand{\R}{\mathbb{R}}
\newcommand{\A}{\mathcal{A}}
\newcommand{\bds}{\boldsymbol}

\newcommand{\best}[1]{{\textcolor{blue}{\textbf{#1}}}}
\newcommand{\second}[1]{{\textcolor{violet}{\textbf{#1}}}}

\title{Principled Data Generation for MetaBBO via Active Task Selection}

\author{Rong-Xi Tan\textsuperscript{1,2}, Jia-Ting Li\textsuperscript{1,2}, Ming Chen\textsuperscript{1,2}, Ke Xue\textsuperscript{1,2}\thanks{Corresponding to Ke Xue and Chao Qian: \texttt{\{xuek,qianc\}@lamda.nju.edu.cn}.}\phantom{$^*$}\ \& Chao Qian\textsuperscript{1,2}\footnotemark[1] \\
\textsuperscript{1}State Key Laboratory of Novel Software Technology, Nanjing University, China\\
\textsuperscript{2}School of Artificial Intelligence, Nanjing University, China
}

\iclrfinalcopy
\begin{document}

\maketitle

\begin{abstract}
Meta-Black-Box Optimization (MetaBBO) aims to acquire generalized optimization strategies by training on extensive sets of functions. While traditional approaches predominantly rely on fixed benchmarks, recent efforts have shifted towards maximizing landscape diversity to broaden training distribution coverage. However, we identify a critical \emph{Diversity-Quality Gap}: Simply increasing landscape variety often yields tasks that are misaligned with the agent's evolving capabilities. To address this, we propose \emph{Hierarchical Active Task Selection (HATS)}, a principled framework that constructs an automated curriculum based on the agent's potential for improvement. HATS introduces a regret-based utility metric measuring the performance gap between the current policy and a reference optimizer to guide a bi-level selection process. Specifically, it combines a multi-armed bandit for dynamic function class weighting with an active sampler for parameter replay. Experiments demonstrate that HATS significantly outperforms diversity-driven baselines and achieves superior generalization on real-world tasks with fewer training steps, highlighting the value of quality-based data generation in MetaBBO.
\end{abstract}

\section{Introduction}
\label{sec:intro}

Black-Box Optimization (BBO), which involves optimizing objective functions where analytical forms and gradient information are inaccessible, is a crucial challenge in real-world complex optimization problem, ranging from hyperparameter tuning~\citep{hpob}, molecule discovery~\citep{shin2025offline} and chip placement~\citep{wiremask-bbo, bboplace-bench}.
Driven by the need to automate these processes, the field of BBO is undergoing a paradigm shift from manually designed heuristics to data-driven automation~\citep{position-fundational,metabbo-survey}. 
Meta-Black-Box Optimization (MetaBBO) aims to learn generalized optimization strategies by training parameterized policies on extensive distributions of problem instances~\citep{metabox-v1,metabox-v2}. 
Unlike traditional BBO optimizers defined by static rules, MetaBBO agents learn search patterns through interactions with training functions, making their performance heavily dependent on the quality and structure of the training distribution~\citep{diverse-bbo}.

To prevent overfitting to static instances, current approaches in MetaBBO predominantly generate data via open-loop diversity maximization. 
Domain Randomization (DR)~\citep{DR}, a technique of randomly sampling function instances, is often employed to augment the training set~\citep{meta-es,chen2024symbol,bbob}.
A recent pioneering work, Diverse-BBO~\citep{diverse-bbo}, incorporates Exploratory Landscape Analysis (ELA) features~\citep{ela} to generate a set of diverse functions. 
These approaches assume that a training set spanning a wide range of landscape features guarantees a robust learned policy. 
However, in this paper, we challenge this assumption.
Through a quantitative analysis of instance quality, we identify a Diversity-Quality Gap that simply maximizing landscape diversity is misaligned with the agent's capabilities. 
This observation underscores the necessity of a principled data generation framework for MetaBBO that accounts for the utility of each function instance for the current agent.

\begin{figure}[t]
    \centering

    \begin{subfigure}[t]{0.43\linewidth}
        \centering
        \makebox[0.07\linewidth][l]{(a)}%
        \includegraphics[
            width=0.92\linewidth,
            valign=t
        ]{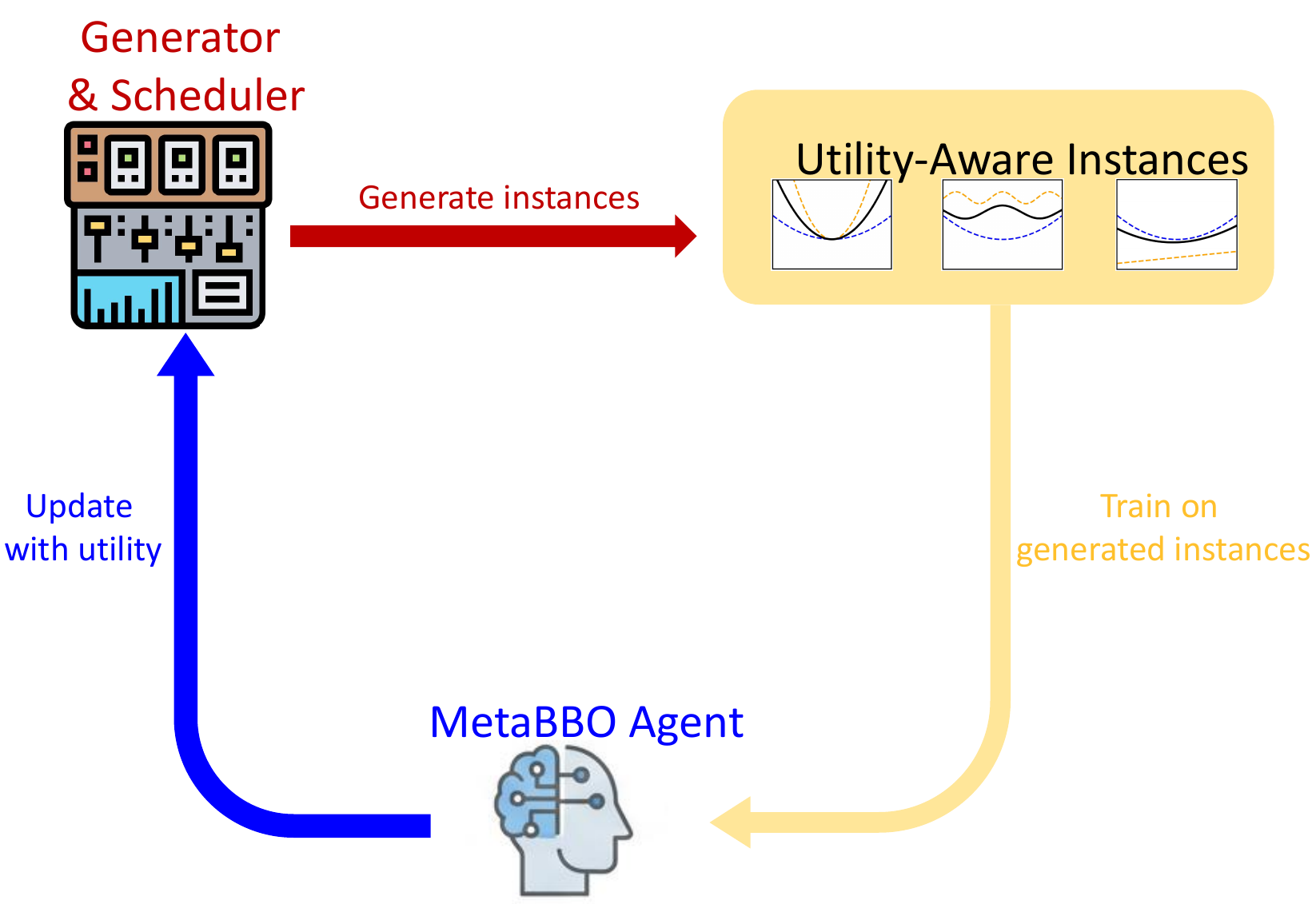}
        \label{fig:framework-overview-a}
    \end{subfigure}
    \hfill
    \begin{subfigure}[t]{0.47\linewidth}
        \centering
        \makebox[0.07\linewidth][l]{(b)}%
        \includegraphics[
            width=0.92\linewidth,
            valign=t
        ]{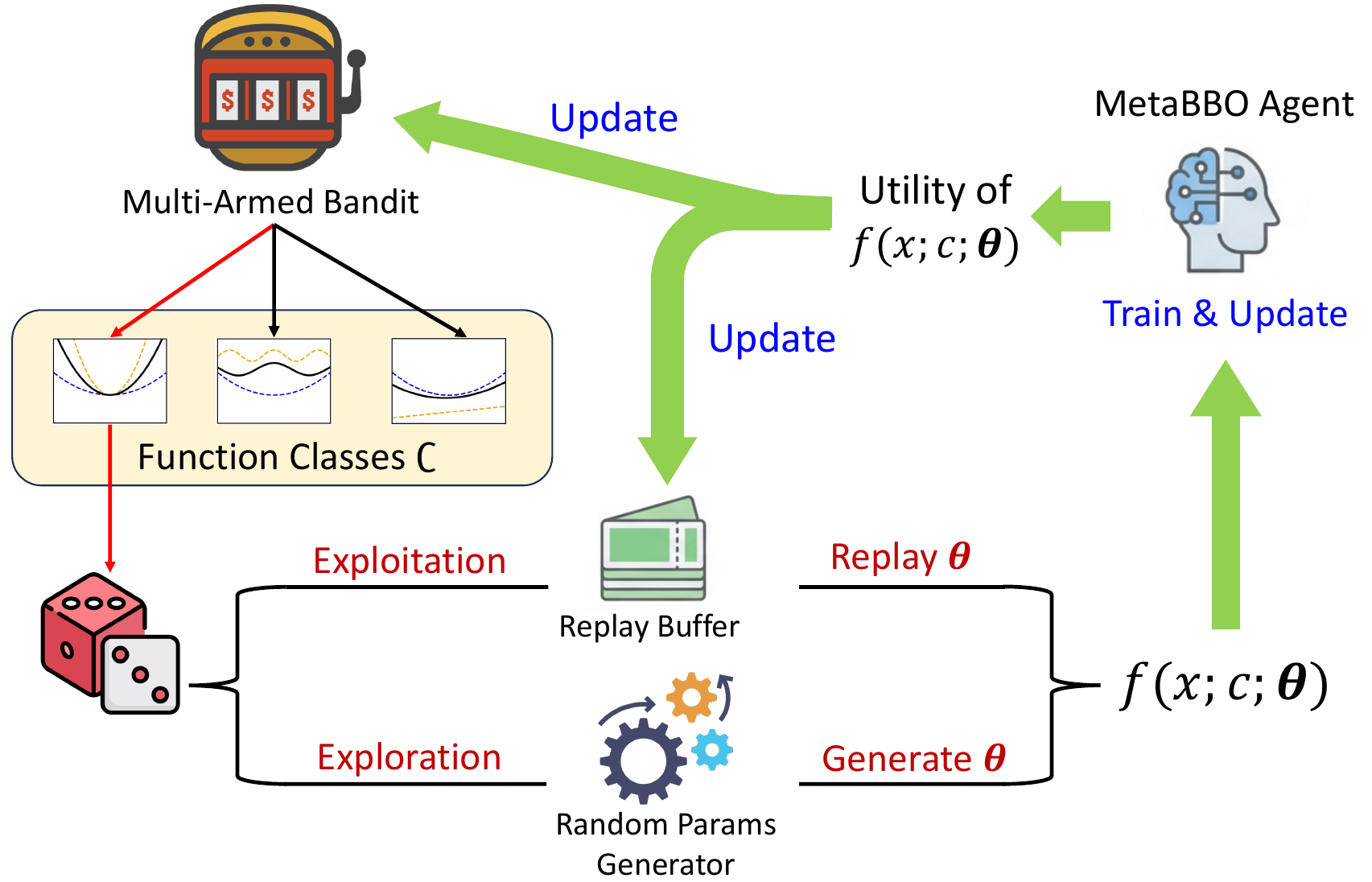}
        \label{fig:framework-overview-b}
    \end{subfigure}

    \caption{
    (a) Illustration of our proposed utility-aware framework for MetaBBO, where the generator
    module interacts with the agent by generating utility-aware instances for agent training
    and receiving utility feedback from the agent.
    (b) Illustration of our proposed HATS framework.
    }
    \label{fig:framework-overview}
\end{figure}

To address this, we propose defining task utility by the agent's dynamic potential for improvement. 
Recognizing that global optima are often intractable in generated functions, we introduce a regret-based utility metric that measures the performance gap between the current MetaBBO policy and a reference optimizer (e.g., MadDE~\citep{madde}). 
This metric serves as a precise learning signal, guiding the generator to construct a curriculum of tasks that are challenging yet solvable relative to the reference.
Using the utility metric, we reformulate the data generation framework for MetaBBO by scheduling instances based on their utilities, as shown in Figure~\ref{fig:framework-overview}(a), and instantiate it as the bi-level HATS framework in Figure~\ref{fig:framework-overview}(b).
This framework views the generation of instances as a closed-loop process, thereby enabling automated curriculum learning for MetaBBO.

Practically, we instantiate this principle as Hierarchical Active Task Selection (HATS), a framework that leverages the inherent structure of BBO problems. 
HATS treats the instance space as a hierarchy $\mathcal{G}=\langle\mathcal{C},\Theta\rangle$, separating global topology (e.g., funnel structures and multi-modality) from local geometric distortions (e.g., rotation and shifting)~\citep{ela,bbob}. 
Specifically, the framework employs a bi-level active selection strategy:
1) A robust multi-armed bandit dynamically allocates the training budget across function classes to ensure topological coverage and prevent catastrophic forgetting;
2) An active sampler prioritizes distortion parameters that maximize the regret-based utility, ensuring the agent continuously trains on the most informative instances for its current stage~\citep{PLR,robust-plr}.

Empirically, we evaluate HATS with SYMBOL~\citep{chen2024symbol} on MetaBox~\citep{metabox-v1,metabox-v2}, showing consistent gains over diversity-based data generation and adaptive function-class dynamics throughout training. Our contributions are summarized as follows:
\begin{itemize}[leftmargin=1em, labelindent=0em]
    \item We empirically demonstrate the Diversity-Quality Gap in MetaBBO, showing that feature diversity correlates poorly with training efficiency.
    \item We propose HATS, a data generation framework that constructs an adaptive curriculum by actively selecting function class and parameters based on the performance gap towards a reference.
    \item We validate HATS across multiple MetaBBO backbones on unseen instances from benchmark suites (CEC and BBOB) and real-world applications including Protein Docking and UAV path planning, showing significant improvements in generalization over state-of-the-art diversity-driven methods.
\end{itemize}

\section{Preliminaries and Related Works}
\subsection{BBO and MetaBBO}
\textbf{BBO.} 
A standard BBO problem involves minimizing a black-box objective function $f:\Omega\mapsto\R$, where $\Omega\subseteq\R^D$ is the search space of dimension $D$. 
The goal is to find $\x^*=\arg\min_{\x\in\Omega}f(\x)$. 
In BBO, the analytical form and gradient information of $f$ are inaccessible, thus a BBO optimizer $\mathcal{A}$ interacts with $f$ solely through queries, receiving a response $y=f(\x)$ for each query $\x$, constrained by a maximum budget $T$ of function evaluations.

\textbf{MetaBBO.} Leveraging meta-learning techniques~\citep{meta-learning-book, maml}, MetaBBO aims to automate the design of BBO algorithms by learning a policy that generalizes across a distribution of problem instances~\citep{metabbo-survey}, showing superior applicability to real-world complex optimization problems~\citep{dac-for-placement}. 
This process can typically be reformulated by a bi-level optimization problem:
\begin{itemize}[leftmargin=1em, labelindent=0em]
    \item \textbf{Inner level}: The optimizer $\A$, controlled by a parameterized policy $\pi_\phi$ (where $\phi$ denotes the meta-parameters, e.g., weights of a neural network), executes the optimization process on a specific problem instance $f$. 
    This generates a trajectory $\tau_f=\{(\x_t,y_t)\}_{t=1}^T$.
    \item \textbf{Meta level}: 
    The objective of MetaBBO is to find the optimal meta-parameters $\phi^*$ that maximize the expected performance metric $\mathcal{M}$, e.g., negative regret, over a target problem distribution $p(f)$. 
    Formally, it can be written by:
    \begin{equation}\label{eq:metabbo-obj}
    \phi^*=\mathop{\arg\max}_{\phi}~\mathbb{E}_{f\sim p(f)}[\mathcal{M}(\A(\pi_\phi, f))].
    \end{equation}
\end{itemize}
Targeting at different parameterized perspectives, existing MetaBBO methods can be categorized into three lines~\citep{metabbo-survey}. 
The first category is algorithm selection, which maintains a pool of algorithms and fits a selector for optimal algorithms~\citep{rl-das,as-2}.
The second category is algorithm configuration~\citep{ac-1,dac-1}, which dynamically adjusts parameter values for a fixed optimizer~\citep{madac,sadn,configx}.
Due to the efficiency of leveraging offline data~\citep{q-mamba} and interpretability~\citep{chen2024symbol}, algorithm generation recently has emerged as a novel solution for MetaBBO. 
\subsection{Data Generation Issues in MetaBBO}
\label{sec:data-metabbo}

A key limitation of current MetaBBO methods is their reliance on a largely static training distribution $p_{\textrm{train}}(f)$.
Training on fixed benchmark functions or static augmentations can lead to overfitting to specific landscape features, limiting generalization to unseen problems~\citep{learned-opt-scales,meta-es}.
To broaden the training distribution, existing methods commonly apply Domain Randomization (DR)~\citep{DR} through transformations such as random shifts and rotations~\citep{optformer,meta-es,chen2024symbol,song2024ribbo}, or explicitly maximize landscape diversity based on ELA features~\citep{ela,neural-ela}, as in Diverse-BBO~\citep{diverse-bbo}.

However, more diverse problem instances are not necessarily more useful for training the current MetaBBO agent, as we show in Section~\ref{sec:data-efficiency}.
Since many MetaBBO agents are trained with RL, the most useful training tasks may change as the agent's capability evolves~\citep{acl-rl-survey}.
This motivates an adaptive, utility-aware training distribution rather than static diversity maximization.

\subsection{Automated Curriculum Learning}
Automated curriculum learning adjusts training tasks as the agent improves, aiming to improve generalization and sample efficiency~\citep{acl-rl-survey}.
DR~\citep{DR} samples tasks uniformly, which can lead to sample inefficiency; Unsupervised Environment Design (UED) instead formulates curriculum generation as a regret-maximization game~\citep{PAIRED}.
Prioritized Level Replay (PLR)~\citep{PLR} prioritizes levels with high value loss for replay, while later work suggests focusing on levels within the agent's learnable zone~\citep{in-the-zone}.
These ideas motivate HATS to select training tasks based on their utility to the current MetaBBO agent; more background is provided in Appendix~\ref{appendix:acl-background}.

\section{Measuring Data Efficiency for MetaBBO}
\label{sec:data-efficiency}
In this section, we re-examine data efficiency for MetaBBO.
Section~\ref{sec:diversity-not-enough} challenges the common practice of maximizing instance feature diversity and reveals the poor correlation between instance diversity and quality.
To efficiently quantify an instance's quality for an agent, Section~\ref{sec:utility-definition} defines performance gap-based utility, while Section~\ref{sec:instance-structure} discusses the instance structures of BBO functions that motivate a principled data generation pipeline for MetaBBO.
\subsection{Does Higher Data Diversity Lead to Better Training Efficiency?}
\label{sec:diversity-not-enough}

As discussed in Section~\ref{sec:data-metabbo}, recent works prioritize maximizing the feature diversity of training set to prevent overfitting.
These approaches operate under the implicit assumption that a training distribution with sufficient diversity, which spans a broad range of landscape characteristics, inherently guarantees a robust and generalizable MetaBBO policy~\citep{diverse-bbo}.
However, this raises a fundamental yet overlooked question: \textit{Does maximizing feature diversity necessarily lead to better training efficiency for the MetaBBO agents?}
We conduct an illustrative experiment to answer this question by analyzing the correlation between feature diversity and training efficiency. 
 
To quantify the diversity contribution of each individual instance, following~\citep{diverse-bbo}, we first obtain a 2-dimensional representation for each function by extracting the ELA features of each function using the \textit{pflacco} package~\citep{pflacco} and applying them to a pre-trained autoencoder provided by~\citep{diverse-bbo}~\footnote{\url{https://github.com/MetaEvo/Diverse-BBO/blob/main/models/autoencoder_epoch_300.pth}} to derive a hidden representation $\mathbf{z}$. 
Then, we compute the averaged Euclidean distance from each problem instance to its $k$ nearest neighbors in the standardized representation space as the diversity score.
Formally, the diversity score $D_i$ for the $i$-th problem instance is calculated by:
\begin{equation}\label{eq:diversity}
D_i=\frac{1}{k}\sum_{j\in\mathcal{N}_i}\|\mathbf{z}_i-\mathbf{z}_j\|_2,
\end{equation}
where $\mathbf{z}_i$ is the normalized feature vector of the $i$-th instance, and $\mathcal{N}_i$ represents the set of indices of its $k$ nearest neighbors. 
A higher score $D_i$ indicates that the instance resides in a sparse region of the feature space, thereby contributing more to the overall diversity.

Additionally, to quantify a problem instance $f$'s quality for MetaBBO, we measure its marginal contribution to the learned policy $\pi_\phi$'s generalization capability. 
Given a finite training set $\mathcal{D}_{\text{train}} = \{f_1, \dots, f_N\}$ and a held-out test set $\mathcal{D}_{\text{test}}$, we adopt a Leave-One-Out (LOO) valuation protocol. 
For each instance $f_i \in \mathcal{D}_{\text{train}}$, we train a MetaBBO agent on the subset $\mathcal{D}_{-i}=\mathcal{D}_{\text{train}} \setminus \{f_i\}$ to obtain the marginal optimal meta-parameters $\phi^*_{-i}$.
The quality score $Q_i$ is defined as the performance gap on $\mathcal{D}_{\text{test}}$ between the policy trained on the full dataset and the policy trained without $f_i$:
\begin{equation}\label{eq:loo-quality}
    Q_i = \mathbb{E}_{f \sim \mathcal{D}_{\text{test}}} \left[ \mathcal{M}(\mathcal{A}(\pi_{\phi^*_{\text{all}}}, f)) \right] - \mathbb{E}_{f \sim \mathcal{D}_{\text{test}}} \left[ \mathcal{M}(\mathcal{A}(\pi_{\phi^*_{-i}}, f)) \right],
\end{equation}
where $\phi^*_{\text{all}}$ denotes the optimal meta-parameters learned from the full training set $\mathcal{D}_{\text{train}}$. 
A higher value of $Q_i$ indicates that instance $f_i$ contains critical features that are essential for the optimizer $\mathcal{A}$ to generalize to unseen problems, thereby signifying high quality.

In practice, we use the function instances set constructed by~\citep{diverse-bbo}~\footnote{\url{https://github.com/MetaEvo/Diverse-BBO/blob/main/save_gp_functions/all_256_programs.pickle}}, which contains 256 diverse instances and encompasses a wide coverage of the ELA features, to train SYMBOL~\citep{chen2024symbol}.
We evaluate the model on the held-out test set from~\citep{diverse-bbo}, which consists of unseen synthetic functions and realistic optimization problems including HPO-B~\citep{hpob}, protein~\citep{protein} and UAV~\citep{uav}. 

\begin{figure}[t]
    \centering

    \begin{subfigure}[t]{0.43\linewidth}
        \centering
        \makebox[0.07\linewidth][l]{(a)}%
        \includegraphics[
            width=0.92\linewidth,
            valign=t
        ]{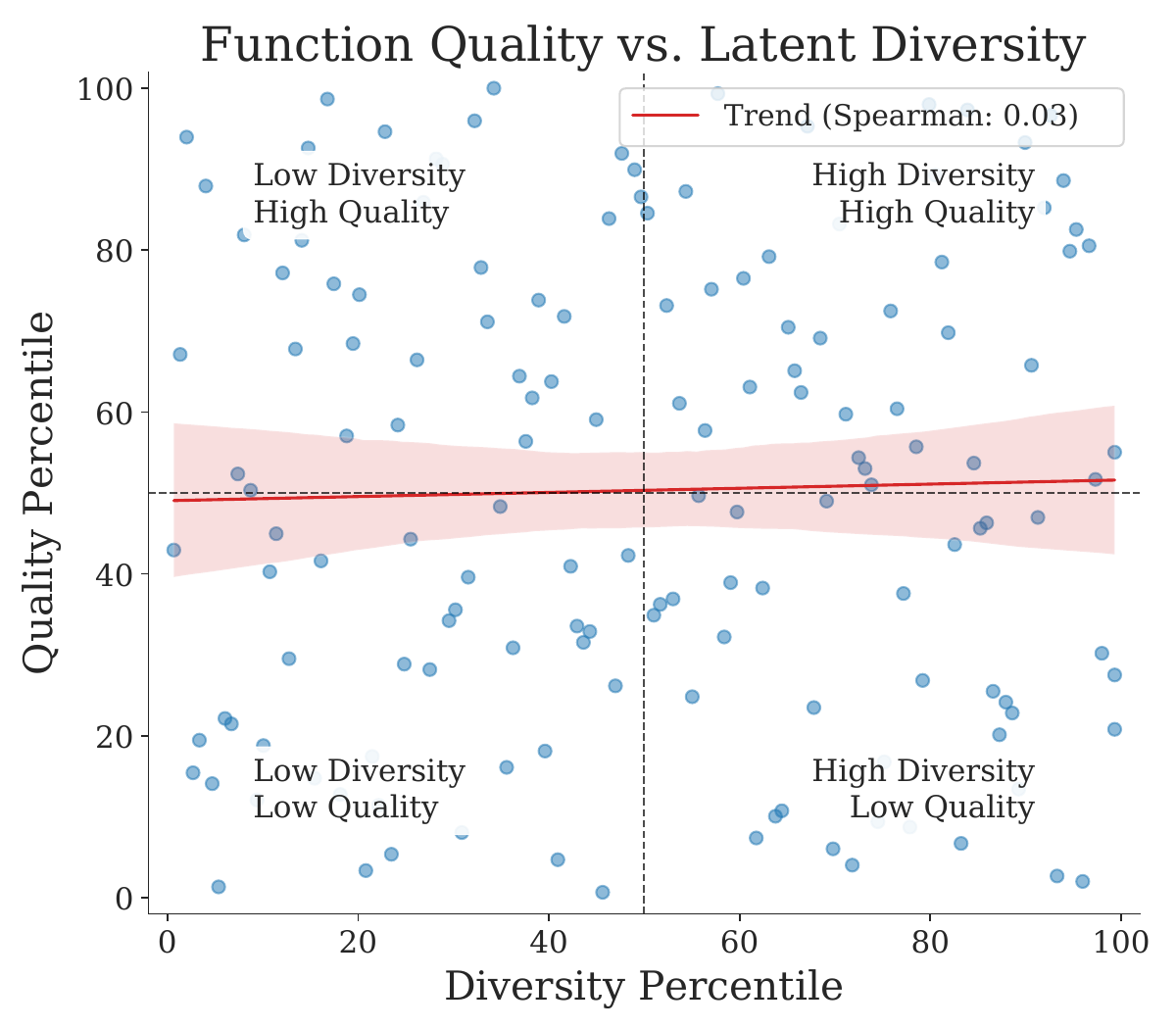}
        \label{fig:data-motivation-a}
    \end{subfigure}
    \hfill
    \begin{subfigure}[t]{0.46\linewidth}
        \centering
        \makebox[0.07\linewidth][l]{(b)}%
        \includegraphics[
            width=0.92\linewidth,
            valign=t,
            trim=0 5mm 0 3mm,
            clip
        ]{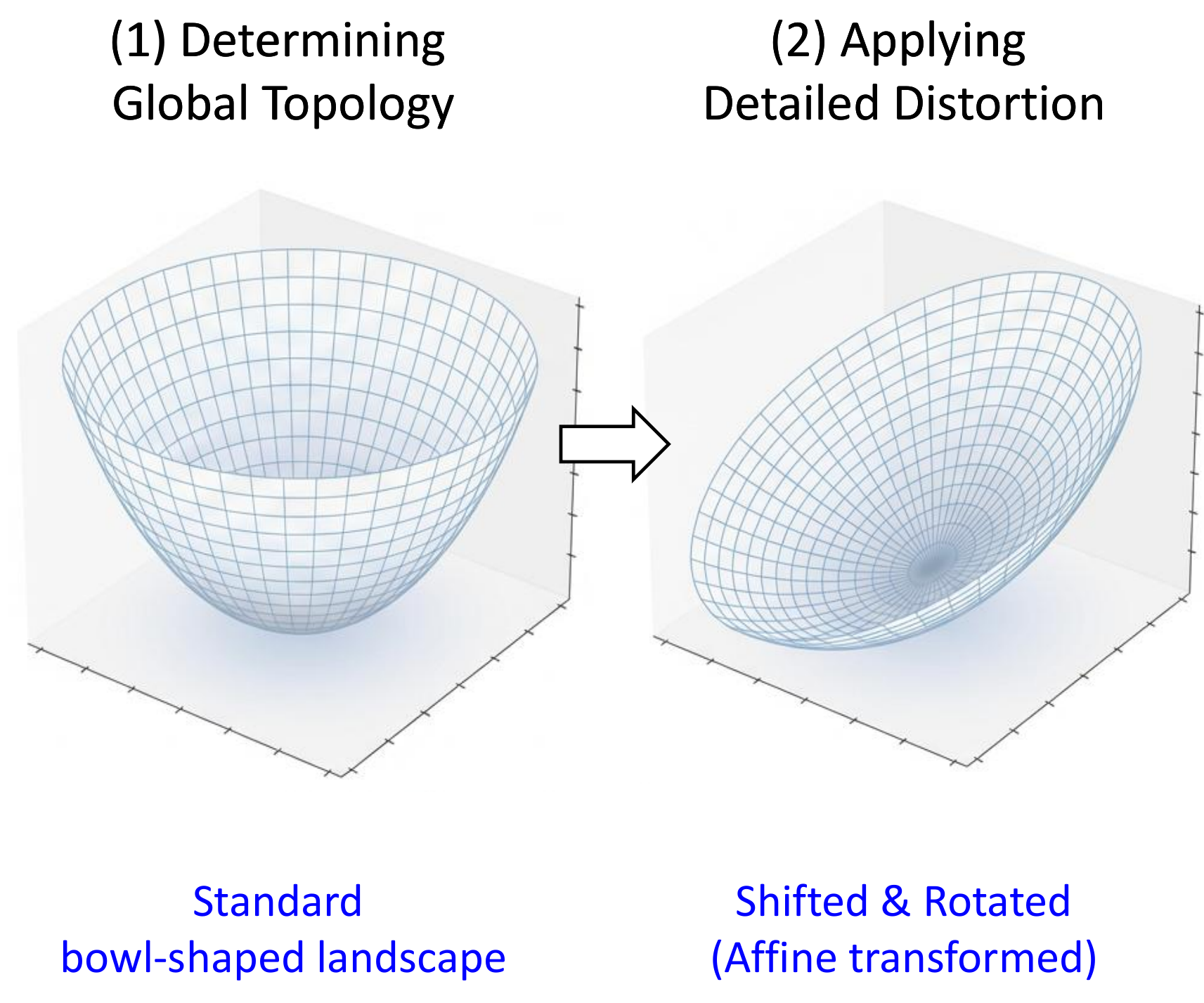}
        \label{fig:data-motivation-b}
    \end{subfigure}

    \caption{
    (a) Scatter plot showing the correlation between function quality and its latent diversity.
    Each point corresponds to one function instance, with diversity measured in the latent space of ELA features and quality quantified by its LOO contribution.
    The near-zero Spearman correlation (0.03) indicates a weak relationship between function quality and diversity.
    (b) Illustration of the hierarchical generation of a bowl-shaped instance.
    }
    \label{fig:data-motivation}
\end{figure}

In Figure~\ref{fig:data-motivation}(a), we visualize the scatter plots of the instances' diversity and quality and calculate their Spearman correlation coefficient, where the data points are uniformly distributed across the entire space, occupying all four quadrants. 
Besides, the extremely low Spearman correlation (0.03) reveals a decoupled nature between these two metrics.
This observation underscores the necessity of explicitly incorporating quality-aware guidance into the training instances generation, rather than relying solely on diversity maximization, motivating us to propose a novel quality-oriented data generation framework for MetaBBO. 

\subsection{Performance Gap-based Utility of Function Instances for MetaBBO}
\label{sec:utility-definition}

The empirical analysis in Section~\ref{sec:diversity-not-enough} establishes that high feature diversity does not guarantee high training quality ($Q_i$). 
Ideally, an effective data generator should prioritize instances with high $Q_i$. 
However, computing the marginal contribution $Q_i$ via the LOO protocol is computationally prohibitive during the training process, as it requires retraining the model from scratch for every candidate instance.
To make quality-aware generation feasible in online MetaBBO agent training, we replace this expensive global LOO objective with a simple proxy: prioritizing instances that have high utility for the current agent, following common practices in RL curriculum design~\citep{acl-rl-survey,bengio2009curriculum,schaul2015prioritized}.

In general RL, training-environment utility is typically characterized by \textit{regret}~\citep{PLR,PAIRED}, which measures the shortfall of the agent's policy against a theoretical optimal policy. 
However, in BBO, relying on the theoretical global optimum is flawed. 
Effective optimization requires underlying landscape structure~\citep{fitness-distance-corr, ela}; yet, randomly generated functions often exhibit chaotic or deceptive landscapes, effectively acting as unoptimizable noise where the global optimum is practically indistinguishable from random chance~\citep{nfl1}. 
In such intractable scenarios, a large global regret reflects the problem's inherent difficulty rather than the agent's incompetence. 
Consequently, defining utility by the global optimum creates a misalignment, encouraging the agent to chase unsolvable noise.

To mitigate the effects of problem difficulty, we instead assess the agent's \textit{relative potential for improvement} against a competent reference optimizer (e.g., CMA-ES~\citep{cmaes} or MadDE~\citep{madde}).
Intuitively, an instance is most informative when the agent underperforms compared to the reference, indicating that the instance is solvable but not yet mastered by the current MetaBBO policy $\pi_\phi$.
Formally, we define the utility as the \textit{Regret Gap} for MetaBBO by:
\begin{equation}\label{eq:gap}
    U(f, \phi) = \log \mathcal{M}(\A_{\text{ref}}(f)) - \log \mathcal{M}(\A(\pi_\phi, f)).
\end{equation}
This gap serves as a precise learning signal. 
A near-zero or negative $U$ suggests the task is already learned (low utility), whereas an excessively large $U$ implies the task might be too difficult with respect to the agent. 
In practice, we employ the concept of ``zone of proximal development"~\citep{ZPD, in-the-zone} to prioritize instances within a constructable range $[u_{\min}, u_{\max}]$. 
This reformulation casts data generation as a \textit{dynamic curriculum learning} problem: Rather than training on a static or random sequence of functions, we construct a curriculum that balances the exploration of high-utility new tasks with the replay of critical historical instances to prevent forgetting.

To facilitate this learning process, a well-constructed structure for BBO functions is required, since navigating the unstructured space of all possible mathematical functions is inefficient.
In the next subsection, we will introduce a structured parameterization that allows us to effectively select and schedule candidate instances.

\subsection{Instance Structures of BBO Functions}
\label{sec:instance-structure}

To construct a structured curriculum, we organize BBO instances by the landscape properties that determine their optimization difficulty.
Following prior landscape analysis~\citep{bbob,ela}, these properties can be broadly divided into two factors.
\textit{Global topology} describes the overall shape of the objective landscape, such as convexity, multimodality, and funnel structure, and determines the type of search behavior required~\citep{ela,global-structure}.
In contrast, \textit{geometric distortion} modifies a fixed topology through transformations such as shifting, rotation, scaling, and conditioning, testing an optimizer's invariance and adaptation~\citep{bbob,oss_vizier}.
Hence, an effective curriculum should cover both different topologies and different distortions within each topology.

Based on this distinction, we model the BBO instance space as a hierarchical generative space
$\mathcal{G}=\langle\mathcal{C},\Theta\rangle$.
The discrete class space $\mathcal{C}=\{c_1,\ldots,c_K\}$ contains base function classes (e.g., Sphere, Rastrigin, and Ackley), where the choice of $c$ determines the global topology.
Given a class $c$, the continuous parameter space $\Theta$ controls its geometric distortion; for example, $\bds{\theta}\in\Theta$ may specify rotation, scaling, and shifting.
As illustrated in Figure~\ref{fig:data-motivation}(b), a concrete instance is generated by first selecting a base class and then applying a geometric transformation:
\begin{equation}\label{eq:hier-gen}
f(\mathbf{x};c;\bds{\theta})
=
g_c\bigl(\mathcal{T}(\mathbf{x};\bds{\theta})\bigr),
\end{equation}
where $g_c$ denotes the canonical base function and $\mathcal{T}$ the transformation parameterized by $\bds{\theta}$.
This hierarchy naturally motivates HATS to schedule function classes and instance parameters separately, as introduced in the next section.

\section{An Efficient Data Generation Framework via Hierarchical Active Task Selection}
\label{sec:method}

In this section, we present our framework, Hierarchical Active Task Selection (HATS), which incorporates a bi-level strategy motivated based on the hierarchical generative space discussed in Section~\ref{sec:instance-structure}.
As shown in Figure~\ref{fig:framework-overview}(b), HATS comprises: 1) a bandit-based mechanism to allocate attention across different function classes $\mathcal{C}$, and 2) an active selection mechanism combining exploration and replay exploitation to schedule the parameters $\bds{\theta}$ efficiently.
One key characteristic of HATS is that the generator is updated \emph{online} using the agent's current competence, rather than optimizing a static diversity objective.

\subsection{Multi-Armed Bandit for Function Class Selection}
\label{sec:hats-mab}

HATS first employs a Multi-Armed Bandit (MAB) mechanism to dynamically allocate the training budget across $K$ distinct function classes, where we treat each function class $c_k \in \mathcal{C}$ (where $K=|\mathcal{C}|$) as an arm. 
We maintain a logit vector $\boldsymbol{\ell}^{(t)} = [\ell_1^{(t)}, \dots, \ell_K^{(t)}]$ to track the time-varying utility of each class. 
Unlike standard stationary bandits, the utility here depends on the agent's evolving competence, showing more robust response to the high stochasticity of the feedback signal.

At each training step $t$, we determine the selection probability $p_k^{(t)}$ for each class $c_k$. To prevent starvation of temporarily difficult classes and ensure continuous exploration, we employ a mixture policy that combines a Softmax distribution with uniform exploration. The probability is defined as:
\begin{equation}\label{eq:mab-prob}
    p_k^{(t)} = (1 - K\epsilon) \cdot \frac{\exp(\ell_k^{(t)})}{\sum_{j=1}^{K} \exp(\ell_j^{(t)})} + \epsilon,
\end{equation}
where $\epsilon \in (0, 1/K)$ is a small constant ensuring a positive minimum sampling probability for every class and preventing rarely sampled arms from starving; specifically, we set $\epsilon=0.05$ for CEC training ($K=10$) and $\epsilon=0.02$ for BBOB training ($K=24$). Based on $\boldsymbol{p}^{(t)}$, a batch of task categories is sampled and subsequently populated with concrete instances by the instance generator.

However, empirically, we find that a vanilla MAB may suffer from probability collapse and catastrophic forgetting; detailed analysis is provided in Appendix~\ref{appendix:robust-mab}.
To mitigate this, we employ two techniques, collectively termed Robust MAB:

\begin{itemize}[leftmargin=1em, labelindent=0em]
\item \textbf{Rank-Based Utility Feedback.}
After the agent trains on the sampled batch, we quantify the value of each class via \emph{Rank Utility}. 
For a batch of $N$ tasks, we calculate the utility of each instance $i$ using Equation~\eqref{eq:gap}, rank them within the batch, and assign a normalized score $v_i \in [0, 1]$ based on the ranking. 
The feedback $\bar{v}_k^{(t)}$ for class $c_k$ is the mean utility of all instances belonging to that class in the current batch.

\item \textbf{Robust Update Mechanism.}
To enable a stable update of the logits, we employ a centered decaying update rule, which is given by:
\begin{equation}\label{eq:mab-update}
    \ell_k^{(t+1)} = \text{clip}\left( \gamma \cdot \ell_k^{(t)} + \eta \cdot (\bar{v}_k^{(t)} - 0.5), \; -L, \; L \right).
\end{equation}
Here, we center the feedback to $0$ by subtracting $0.5$, ensuring that classes performing below the median are penalized while those above are rewarded. The decay factor $\gamma \in (0, 1)$ induces a forgetting mechanism for outdated priors, allowing the scheduler to adapt to the agent's new capabilities.
\end{itemize}

\subsection{Active Parameter Sampling via Utility-Aware Replay}
\label{sec:hats-plr}
Once the MAB allocates the training budget to a specific function class $c$, we employ a Utility-Aware Replay (UAR) strategy to determine the specific parameters $\bds{\theta}$.
While building upon the Prioritized Level Replay (PLR) framework~\citep{PLR}, our mechanism is tailored to select instances from a global buffer $\mathcal{B}$ that maximize training efficiency for the chosen class.
The overall procedure is outlined in Algorithm~\ref{alg:robust-plr}.

\begin{algorithm}[t]
    \caption{UAR for Instance Parameters}
    \label{alg:robust-plr}
    \begin{algorithmic}[1]
        \STATE \textbf{Input:} Function class $c$, global buffer $\mathcal{B}$, policy $\pi_\phi$
        \STATE Sample replay decision $d \sim \text{Bernoulli}(\rho)$
        \IF{$d = 0$ \textbf{or} no valid tasks for $c$ in $\mathcal{B}$}
            \STATE \textcolor{gray}{// Exploration: Generate new task, no gradient update}
            \STATE Sample parameters $\bds{\theta}$ from generator for class $c$
            \STATE Collect trajectory $\tau$ on $\bds{\theta}$ with \emph{stop-gradient} $\phi_\perp$
        \ELSE
            \STATE \textcolor{gray}{// Replay: Sample based on Score $S$ (Equation~\eqref{eq:plr_prob})}
            \STATE Sample $\bds{\theta} \sim \mathcal{B}$ restricted to class $c$ using prob. $\bds{p}$
            \STATE Collect trajectory $\tau$ on $\bds{\theta}$ and update $\phi$ with rewards $R(\tau)$
        \ENDIF
        \STATE Compute utility $u = U(\tau, \pi)$ via Equation~\eqref{eq:gap}
        \STATE Update global buffer $\mathcal{B}$ with instance $\bds{\theta}$ and utility $u$
    \end{algorithmic}
\end{algorithm}
\textbf{Prioritized Sampling Strategy.}
To select the most effective training tasks from the global buffer $\mathcal{B}$, we calculate a sampling score $S_i$ for each instance $i$ belonging to class $c$. 
Let $u_i$ denote the utility score derived from Equation~\eqref{eq:gap} and $t_i$ denote the staleness (number of steps since the last replay). The score is computed as:
\begin{equation}\label{eq:plr-score}
    S_i = \frac{u_i - u_{\min}}{u_{\max} - u_{\min}} + \alpha \cdot \frac{t_i - t_{\min}}{t_{\max} - t_{\min}},
\end{equation}
where values are normalized over the valid subset of the buffer. The hyperparameter $\alpha$ balances the trade-off between utility and staleness.
The probability of replaying instance $i$ is then determined by a Softmax distribution with temperature $\tau$:
\begin{equation}
    p_i = \frac{\exp(S_i / \tau)}{\sum_{j \in \mathcal{B}_c} \exp(S_j / \tau)},
    \label{eq:plr_prob}
\end{equation}
where $\mathcal{B}_c$ denotes the subset of the global buffer containing instances of class $c$, and $\tau$ is set to $1.0$. This ensures that the agent focuses on tasks with high training utility while periodically revisiting older instances to prevent catastrophic forgetting~\citep{PLR}.

Additionally, as recommended by~\citep{robust-plr}, UAR employs a \emph{stop-gradient} operation $\phi_\perp$ to exploration trajectories, updating the policy $\pi$ only on verified replay instances.
As detailed in Algorithm~\ref{alg:robust-plr}, when exploring new parameters (Line 4), we generate $\bds{\theta}$ from the function class $c$ but strictly suppress the policy update. 
Gradient updates are performed only during the replay phase (Line 8), ensuring the agent learns only from tasks with high utility.

\section{Experiments}
\label{sec:exp}
We evaluate HATS from multiple perspectives. Section~\ref{sec:exp-setup} describes the experimental setup and tasks, while Section~\ref{sec:exp-results} presents results organized around several Research Questions (RQs).

\subsection{Experimental Setup and Selected Tasks}
\label{sec:exp-setup}
\textbf{MetaBBO Backbones.}
We use SYMBOL~\citep{chen2024symbol} as the primary backbone, and further evaluate
HATS with LDE~\citep{lde}, B2Opt~\citep{b2opt}, and GLHF~\citep{glhf} to examine its applicability across different
MetaBBO agents. For SYMBOL, we replace its original DR-based data generation with HATS while keeping the remaining components unchanged. Details of SYMBOL are provided in Appendix~\ref{appendix:detail-symbol}.

\textbf{Compared Baselines.}
We compare HATS against three categories of baselines:
(1) \textbf{Static training}: Models trained on fixed function sets without augmentation or curriculum mechanisms.
This includes two standard benchmarks, BBOB~\citep{bbob} and CEC~\citep{cec2021}, as well as fixed function sets generated by recent methods, MA-BBOB~\citep{mabbob} and Diverse-BBO~\citep{diverse-bbo};
(2) \textbf{DR}~\citep{DR}: A widely adopted baseline that applies random shifts and rotations to augment the base functions (i.e., functions from BBOB and CEC); and
(3) \textbf{Prioritized replay variants}: To isolate the contributions of reference-based utility and hierarchical scheduling, we further include three compatible PLR-style methods on both BBOB and CEC: \textbf{Flat replay}, which retains our regret-gap utility in Equation~\eqref{eq:gap} with UAR over all instances but removes MAB class-level scheduling; \textbf{PLR$^{\perp}$-L1}, which combines the L1 value-loss $|\hat{R}_t - V(s_t)|$ scoring~\citep{PLR} computed from SYMBOL's native $n$-step return targets with replay-only updates~\citep{robust-plr}; and \textbf{HATS-L1}, which retains our two-level hierarchy (MAB + UAR) while replacing the reference-gap utility with L1 value loss.
Implementation details and HATS hyperparameters can be found in Appendix~\ref{appendix:implementation-details}.

\textbf{Tasks.}
We evaluate all methods on two types of tasks:
unseen synthetic instances constructed from CEC~\citep{cec2021} and BBOB~\citep{bbob} with unseen shifts and rotations, and realistic optimization tasks from MetaBox~\citep{metabox-v1,metabox-v2} for evaluation, including Protein Docking~\citep{protein} and UAV~\citep{uav}~\footnote{We exclude HPO-B~\citep{hpob} tasks from MetaBox due to HPO-B's unstable surrogate evaluation, as demonstrated by~\citep{PFNs4BO}.}.
Specifically, training on CEC and BBOB covers $K=10$ and $K=24$ function classes of $10$D problems, respectively. Held-out synthetic evaluations use fixed seeds on $10$D instances, whereas Protein Docking ($12$D) and UAV ($30$D) use the MetaBox \texttt{difficulty=all} split (see Appendix~\ref{appendix:implementation-details} for full details).
These selected tasks cover a wide range of 1,424 distinct instances to assess all methods' performance.

\begin{table*}[!t]
\centering
\caption{Normalized performance on CEC, BBOB, Protein Docking and UAV using SYMBOL as the MetaBBO backbone, where the best and runner-up on each task are \best{blue} and \second{violet}.
The results are scaled to the range $[0, 1]$, and lower values indicate better performance. 
}
\resizebox{\linewidth}{!}{
\begin{tabular}{ll|cccc|cc}
\toprule
Base Function & Training Strategy & CEC & BBOB & Protein Docking & UAV & Avg. Score & Avg. Rank \\ \midrule

MA-BBOB & Static & 0.245 $\pm$ 0.202 & 0.237 $\pm$ 0.189 & 0.636 $\pm$ 0.108 & 0.432 $\pm$ 0.261 & 0.387 & 12.25 \\ \midrule
Diverse-BBO & Static & 0.302 $\pm$ 0.195 & 0.359 $\pm$ 0.156 & 0.295 $\pm$ 0.071 & 0.484 $\pm$ 0.262 & 0.360 & 11.25 \\ \midrule

\multirow{6}{*}{BBOB} & Static & 0.251 $\pm$ 0.177 & 0.262 $\pm$ 0.134 & 0.395 $\pm$ 0.109 & 0.486 $\pm$ 0.280 & 0.349 & 12.50 \\
 & DR & 0.126 $\pm$ 0.108 & 0.149 $\pm$ 0.107 & 0.163 $\pm$ 0.056 & 0.356 $\pm$ 0.236 & 0.199 & 7.00 \\
 & Flat replay & 0.088 $\pm$ 0.081 & 0.090 $\pm$ 0.086 & 0.415 $\pm$ 0.094 & 0.367 $\pm$ 0.218 & 0.240 & 7.25 \\
 & PLR$^{\perp}$-L1 & 0.118 $\pm$ 0.097 & 0.113 $\pm$ 0.079 & 0.367 $\pm$ 0.087 & 0.226 $\pm$ 0.193 & 0.206 & 5.50 \\
 & HATS-L1 & 0.228 $\pm$ 0.184 & 0.215 $\pm$ 0.153 & 0.762 $\pm$ 0.120 & 0.239 $\pm$ 0.250 & 0.361 & 10.00 \\
 & \textbf{HATS (Ours)} & 0.076 $\pm$ 0.075 & 0.110 $\pm$ 0.090 & 0.349 $\pm$ 0.084 & \second{0.219 $\pm$ 0.214} & \second{0.189} & \second{3.75} \\ \midrule

\multirow{6}{*}{CEC} & Static & 0.181 $\pm$ 0.132 & 0.175 $\pm$ 0.114 & 0.513 $\pm$ 0.123 & 0.308 $\pm$ 0.253 & 0.294 & 10.00 \\
 & DR & 0.125 $\pm$ 0.108 & 0.162 $\pm$ 0.133 & \best{0.093 $\pm$ 0.048} & 0.276 $\pm$ 0.270 & 0.164 & 5.75 \\
 & Flat replay & 0.130 $\pm$ 0.114 & 0.160 $\pm$ 0.119 & 0.383 $\pm$ 0.091 & 0.328 $\pm$ 0.211 & 0.250 & 8.75 \\
 & PLR$^{\perp}$-L1 & 0.081 $\pm$ 0.084 & 0.127 $\pm$ 0.109 & 0.360 $\pm$ 0.096 & 0.261 $\pm$ 0.194 & 0.207 & 5.50 \\
 & HATS-L1 & 0.070 $\pm$ 0.073 & \second{0.097 $\pm$ 0.087} & 0.345 $\pm$ 0.079 & 0.285 $\pm$ 0.203 & 0.199 & 4.25 \\
 & \textbf{HATS (Ours)} & \best{0.043 $\pm$ 0.051} & \best{0.064 $\pm$ 0.059} & \second{0.095 $\pm$ 0.050} & \best{0.102 $\pm$ 0.225} & \best{0.076} & \best{1.25} \\

\bottomrule
\end{tabular}}
\label{tab:main-results}
\end{table*}

\subsection{Experiment Results}
\label{sec:exp-results}
In this section, we present our experimental findings, aiming to answer the following RQs.

\textbf{RQ1: Does HATS improve generalization of the learned MetaBBO agent?}
Table~\ref{tab:main-results} shows that:
(1) HATS achieves the best overall performance within both training groups, obtaining average scores of $0.189$ and $0.076$ with average ranks of $3.75$ and $1.25$ on BBOB and CEC, respectively, outperforming DR across nearly all tasks.
(2) DR consistently outperforms static training, demonstrating the benefit of instance augmentation.
(3) Comparing against compatible prioritized replay baselines, HATS consistently surpasses Flat replay, PLR$^{\perp}$-L1, and HATS-L1 on both BBOB and CEC. Removing function-class scheduling degrades out-of-distribution generalization on real-world tasks, while replacing the reference-based gap with critic value-loss error yields weaker learning signals, confirming the joint benefits of our regret-gap utility and hierarchical scheduling.

\begin{figure}[t]
    \centering
    \includegraphics[width=0.47\linewidth]{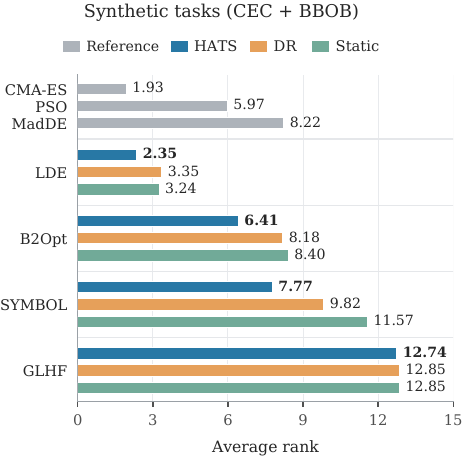}
    \hfill
    \includegraphics[width=0.47\linewidth]{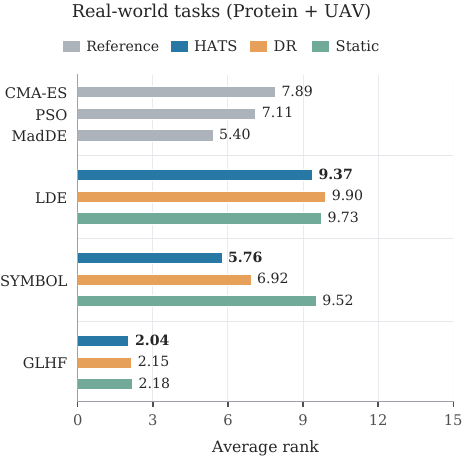}
    \caption{
    Average rank of HATS, DR, and static training across MetaBBO backbones on synthetic (left) and real-world tasks (right; lower is better).
    Gray bars denote reference baselines.
    B2Opt is omitted from real-world tasks because it is a dimension-dependent
    model trained under 10D, whereas Protein Docking and UAV are not 10D problems.
    }
    \label{fig:backbone_generalization}
\end{figure}

\textbf{RQ2: Does HATS generalize across different MetaBBO backbones?}
We further evaluate HATS with LDE, B2Opt, and GLHF in addition to SYMBOL. Figure~\ref{fig:backbone_generalization} shows that HATS achieves lower average ranks than DR and static training in all seven evaluated backbone-task settings, demonstrating broad applicability across MetaBBO agents. Moreover, under matched total training evaluation budgets that account for both learner and reference evaluations, HATS still consistently outperforms DR and static training on teacher-free backbones (Appendix~\ref{appendix:matched-budget}).

\begin{figure}[t]
    \centering

    \begin{minipage}[b]{0.45\linewidth}
        \centering
        \makebox[0.07\linewidth][l]{%
            \raisebox{3.175cm}[0pt][0pt]{(a)}%
        }%
        \includegraphics[
            width=0.92\linewidth,
            valign=b
        ]{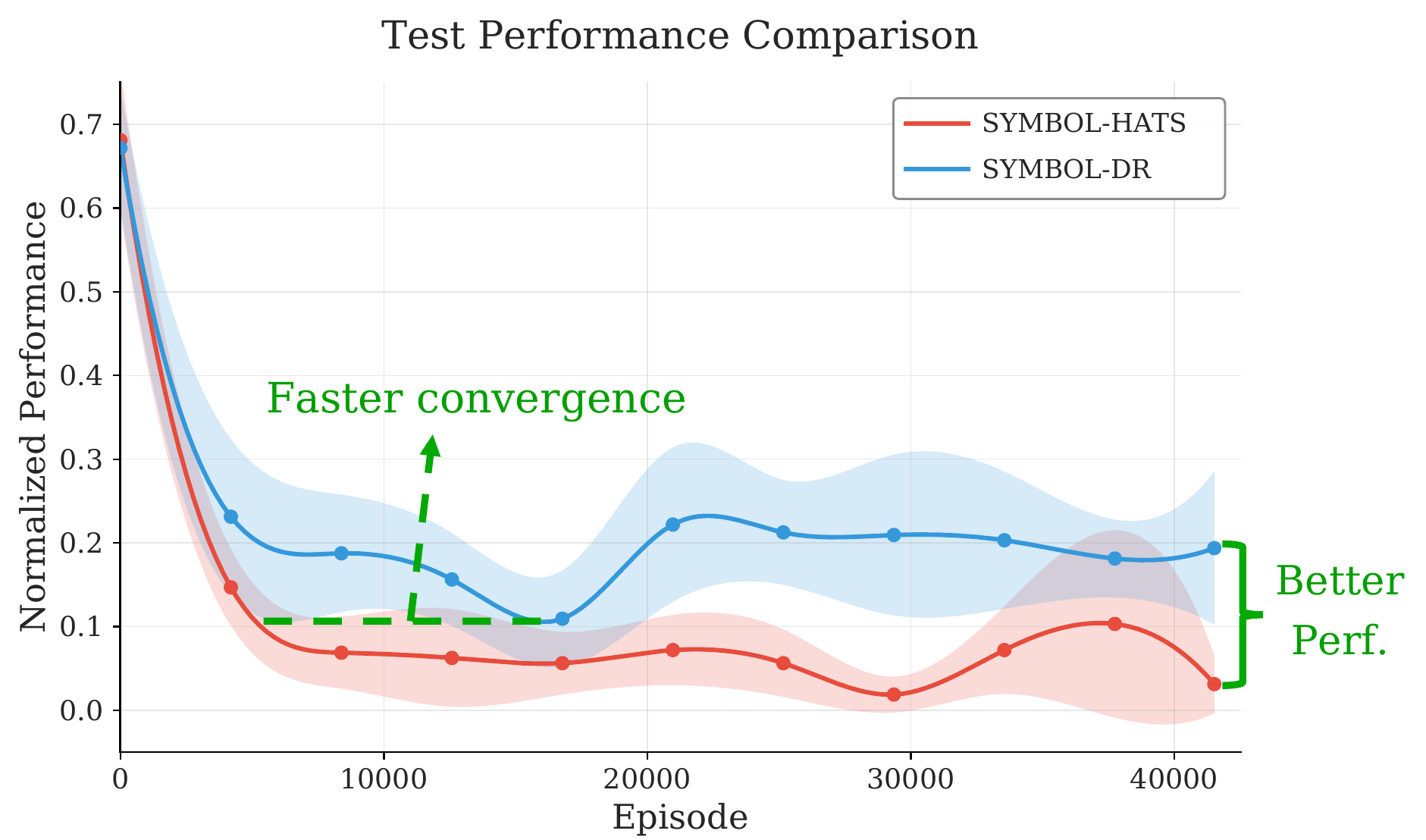}
    \end{minipage}
    \hfill
    \begin{minipage}[b]{0.49\linewidth}
        \centering
        \makebox[0.07\linewidth][l]{%
            \raisebox{3.175cm}[0pt][0pt]{(b)}%
        }%
        \includegraphics[
            width=0.92\linewidth,
            valign=b
        ]{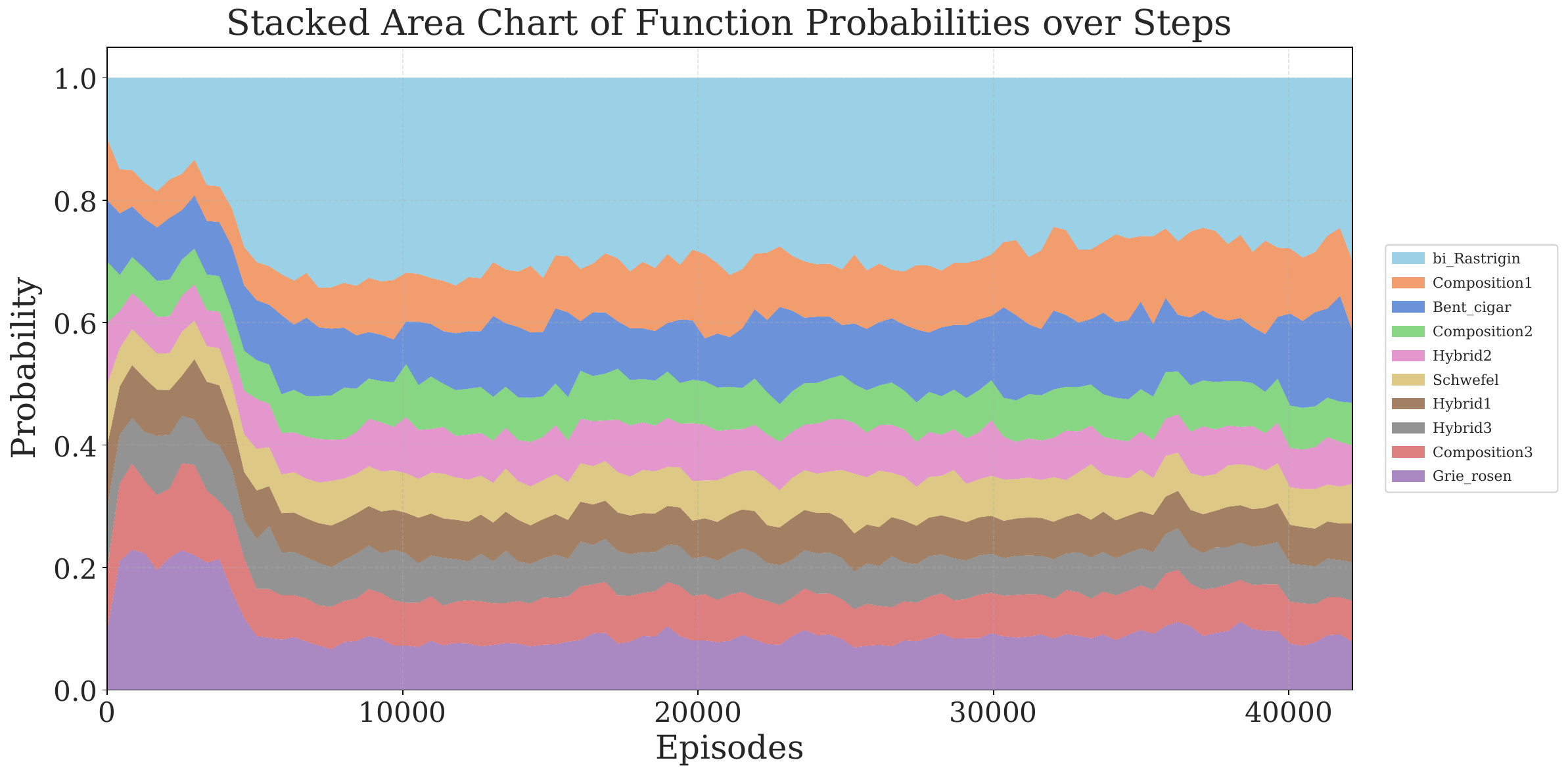}
    \end{minipage}

    \caption{
    (a) Normalized performance (minimization) comparison between SYMBOL-HATS (red)
    and SYMBOL-DR (blue).
    (b) Stacked area chart of function class probability distribution of CEC functions
    during training.
    }
    \label{fig:score-dynamics}
\end{figure}

\textbf{RQ3: Does HATS improve training efficiency?}
Figure~\ref{fig:score-dynamics} (a) shows that SYMBOL combined with HATS converges faster and achieves better final performance than that with DR on CEC.
In particular, HATS reaches a low normalized test performance within fewer training episodes and maintains lower test performance than DR throughout training.
We also provide a case study of these two agents in Appendix~\ref{appendix:case-study}.

\textbf{RQ4: How does the curriculum evolve under HATS?}
Figure~\ref{fig:score-dynamics} (b) shows that HATS dynamically adapts the sampling distribution over CEC function classes during training.
The probability of the relatively simple \texttt{Grie\_rosen} first increases and then decreases after about 5000 episodes, suggesting that the agent has gradually learned to handle this class and therefore requires less training on it.
Meanwhile, sampling shifts toward the harder multimodal \texttt{bi\_Rastrigin}, showing that HATS changes its training focus as the agent's capability evolves.

\begin{figure}[t]
    \centering
    \includegraphics[width=0.325\linewidth]{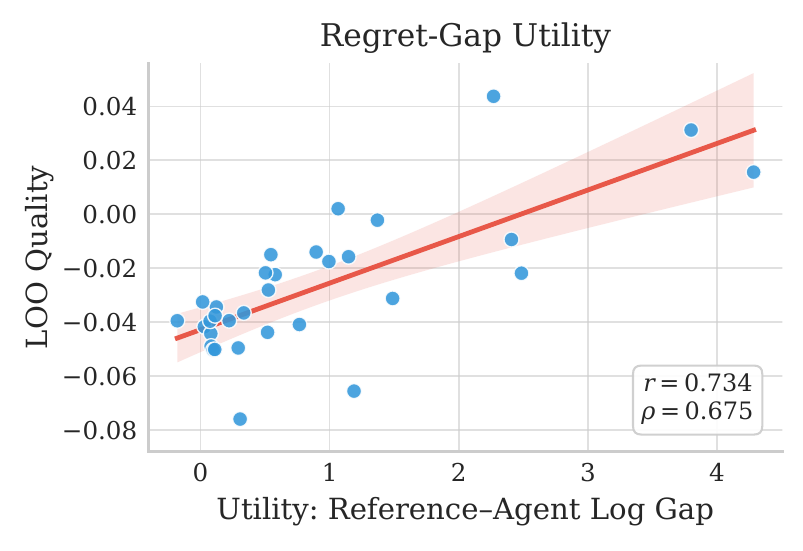}\hfill
    \includegraphics[width=0.325\linewidth]{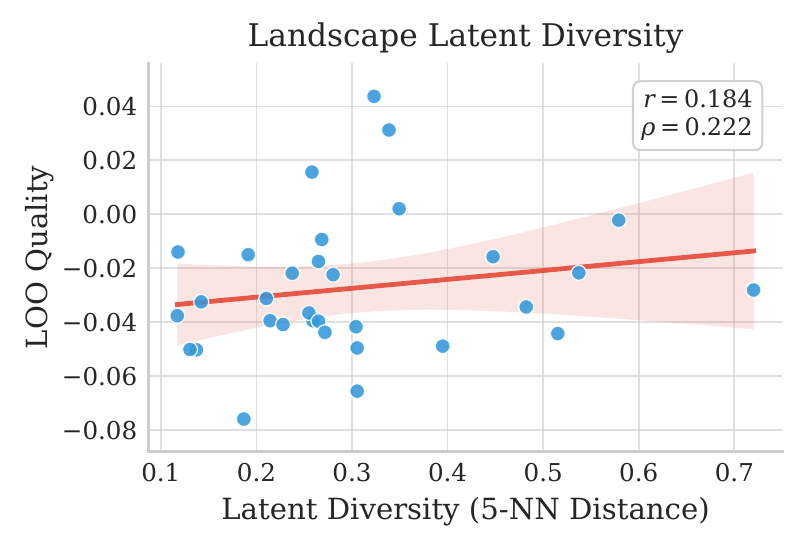}\hfill
    \includegraphics[width=0.325\linewidth]{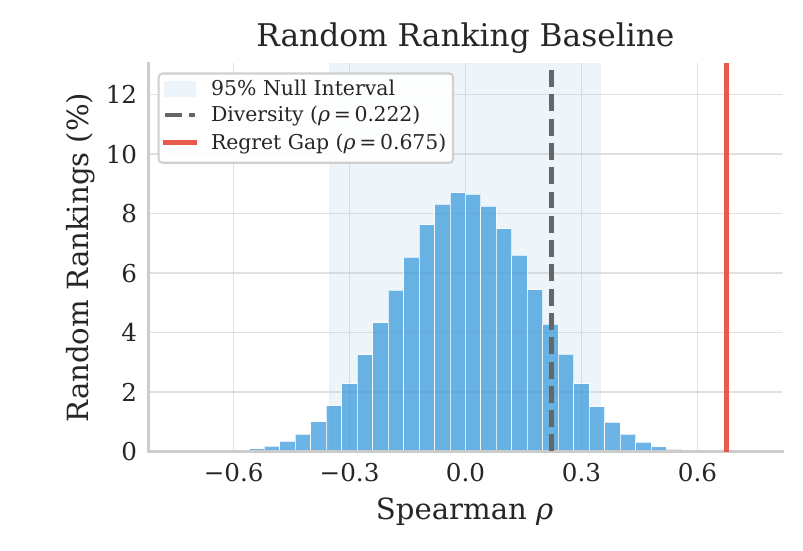}
    \caption{Correlation between task scoring metrics and LOO training quality over $32$ generated instances from a mid-training checkpoint: (left) our reference--agent log-gap utility ($r=0.734$, $\rho=0.675$), (middle) latent diversity ($r=0.184$, $\rho=0.222$), and (right) Spearman $\rho$ distribution of $200{,}000$ random rankings ($95\%$ null interval shaded).}
    \label{fig:loo-utility-validation}
\end{figure}

\textbf{RQ5: Does the proposed regret-gap utility predict actual learning gains?}
While Section~\ref{sec:diversity-not-enough} shows that feature diversity correlates poorly with LOO quality, we further verify whether our reference--agent gap in Equation~\eqref{eq:gap} directly predicts learning gains. Starting from a mid-training SYMBOL checkpoint, we randomly generate $32$ instances and compare three scoring signals against LOO quality in Equation~\eqref{eq:loo-quality}: (1) our regret-gap utility in Equation~\eqref{eq:gap}, (2) latent diversity in Equation~\eqref{eq:diversity}, and (3) $200{,}000$ random rankings. As shown in Figure~\ref{fig:loo-utility-validation}, our utility correlates much more strongly with LOO quality ($r=0.734$, $\rho=0.675$) than latent diversity ($r=0.184$, $\rho=0.222$). Its Spearman correlation also exceeds the central $95\%$ random-ranking interval ($[-0.350, 0.349]$), whereas latent diversity falls within it, confirming that the reference--agent gap is an informative proxy for instance quality. Further controlled continuation experiments across utility zones confirm that prioritizing moderate-gap instances consistently yields the largest learning gains, while overly hard tasks can even degrade policy performance (Appendix~\ref{appendix:zone-continuation}).

\textbf{Additional analysis.}
In Appendix~\ref{appendix:zone-continuation},~\ref{appendix:teacher-sensitivity}, and~\ref{appendix:robust-mab}, we validate continuation gains across utility zones, study reference sensitivity and post-surpassing dynamics (showing MadDE provides a well-matched curriculum reference and the agent continues improving after surpassing a weak reference), and analyze Robust MAB.
We ablate MAB and UAR in Appendix~\ref{sec:ablate} to validate each component, and further examine HATS's applicability to few-shot offline BBO with ExPT in Appendix~\ref{appendix:expt-results}.
In Appendix~\ref{appendix:case-study},~\ref{appendix:matched-budget},~\ref{appendix:hierarchy-scoring}, and~\ref{appendix:curve}, we provide a case study of HATS and DR, compare matched budgets, briefly explain the prioritized replay results, and report the optimization curves across tasks, respectively.
\section{Conclusion}
\label{sec:conclusion}
In this work, we focus on the data generation issue for MetaBBO. We identify the Diversity-Quality
Gap in MetaBBO and propose HATS, a framework that bridges this gap by dynamically selecting
a curriculum based on a regret-based utility metric. Across multiple MetaBBO backbones, HATS
achieves lower average ranks than DR and static training on both synthetic and real-world tasks,
while also showing faster convergence on CEC. These results highlight the benefit of adapting training tasks to the agent's evolving capabilities rather than maximizing diversity alone for more effective and efficient MetaBBO training.
Future work may extend HATS to a generator that not only schedules instances but
also creates effective data iteratively, e.g., via generating symbolic functions~\citep{meidani2024snip,wang2025synthetic}, as well as exploring adaptive or co-evolving references.

\bibliography{iclr2027_conference}

@inproceedings{
chen2024symbol,
title={{SYMBOL}: Generating Flexible Black-Box Optimizers through Symbolic Equation Learning},
author={Jiacheng Chen and Zeyuan Ma and Hongshu Guo and Yining Ma and Jie Zhang and Yue-Jiao Gong},
booktitle={Proceedings of the 12th International Conference on Learning Representations},
year={2024},
address={Vienna, Austria}
}

@inproceedings{diverse-bbo,
      title={Instance Generation for Meta-Black-Box Optimization through Latent Space Reverse Engineering}, 
      author={Chen Wang and Yue-Jiao Gong and Zhiguang Cao and Zeyuan Ma},
      year={2026},
  booktitle={Proceedings of the 40th AAAI Conference on Artificial Intelligence},
address={Singapore},
}

@inproceedings{cec2021,
  title={Problem definitions and evaluation criteria for the {CEC} 2021 on Single Objective Bound Constrained Numerical Optimization},
  author={Ali Wagdy Mohamed and Anas A Hadi and Ali Khater Mohamed and Prachi Agrawal and Abhishek Kumar and P. N. Suganthan},
  year={2021},
  booktitle={Proceedings of the IEEE Congress of Evolutionary Computation},
}

@article{bbob,
  author  = {Hansen, Nikolaus and Auger, Anne and Ros, Raymond and Mersmann, Olaf and Tu{\v{s}}ar, Tea and Brockhoff, Dimo},
  title   = {{COCO}: A platform for comparing continuous optimizers in a black-box setting},
  journal = {Optimization Methods and Software},
  volume  = {36},
  number  = {1},
  pages   = {114--144},
  year    = {2021},
}

@article{protein,
  title={Harnessing protein folding neural networks for peptide--protein docking},
  author={Tsaban, Tomer and Varga, Julia K and Avraham, Orly and Ben-Aharon, Ziv and Khramushin, Alisa and Schueler-Furman, Ora},
  journal={Nature Communications},
  volume={13},
  pages={176},
  year={2022},
}

@inproceedings{
hpob,
title={{HPO-B}: A Large-Scale Reproducible Benchmark for Black-Box {HPO} based on {O}pen{ML}},
author={Sebastian Pineda Arango and Hadi Samer Jomaa and Martin Wistuba and Josif Grabocka},
booktitle={Advances in Neural Information Processing Systems, Datasets and Benchmarks Track},
year={2021},
address={Virtual}
}

@article{uav,
      title={Benchmarking global optimization techniques for unmanned aerial vehicle path planning}, 
      author={Mhd Ali Shehadeh and Jakub Kudela},
      year={2025},
      journal={arxiv:2501.14503},
}

@ARTICLE{metabbo-survey,
  author={Ma, Zeyuan and Guo, Hongshu and Gong, Yue-Jiao and Zhang, Jun and Tan, Kay Chen},
  journal={IEEE Transactions on Evolutionary Computation}, 
  title={Toward Automated Algorithm Design: A Survey and Practical Guide to Meta-Black-Box-Optimization}, 
  year={2025},
  volume={},
  number={},
  pages={1-1},}

@inproceedings{dac-for-placement,
  author    = {Lu, Chen and Xue, Ke and Chen, Ruo-Tong and Shi, Yunqi and Xu, Siyuan and Yuan, Mingxuan and Qian, Chao and Zhou, Zhi-Hua},
  title     = {Dynamic Algorithm Configuration for Global Placement},
  booktitle = {Proceedings of the 2026 Design, Automation \& Test in Europe Conference \& Exhibition},
  year      = {2026},
  address   = {Verona, Italy},
}

@incollection{meta-learning-book,
  author       = {Sebastian Thrun and
                  Lorien Y. Pratt},
  title        = {Learning to Learn: Introduction and Overview},
  booktitle    = {Learning to Learn},
  pages        = {3--17},
  publisher    = {Springer},
  year         = {1998},
}

@inproceedings{maml,
author = {Finn, Chelsea and Abbeel, Pieter and Levine, Sergey},
title = {Model-agnostic meta-learning for fast adaptation of deep networks},
year = {2017},
booktitle = {Proceedings of the 34th International Conference on Machine Learning},
pages = {1126--1135},
address= {Sydney, Australia},
}

@inproceedings{learned-opt-scales,
author = {Wichrowska, Olga and Maheswaranathan, Niru and Hoffman, Matthew W. and Colmenarejo, Sergio G\'{o}mez and Denil, Misha and de Freitas, Nando and Sohl-Dickstein, Jascha},
title = {Learned optimizers that scale and generalize},
year = {2017},
booktitle = {Proceedings of the 34th International Conference on Machine Learning},
pages = {3751--3760},
address = {Sydney, Australia},
}

@inproceedings{acl-rl-survey,
  title     = {Automatic Curriculum Learning For Deep {RL}: A Short Survey},
  author    = {Portelas, R{\'e}my and Colas, C{\'e}dric and Weng, Lilian and Hofmann, Katja and Oudeyer, Pierre-Yves},
  booktitle = {Proceedings of the 29th International Joint Conference on Artificial Intelligence},
  pages     = {4819--4825},
  year      = {2020},
}

@inproceedings{
meta-es,
title={Discovering Evolution Strategies via Meta-Black-Box Optimization},
author={Robert Tjarko Lange and Tom Schaul and Yutian Chen and Tom Zahavy and Valentin Dalibard and Chris Lu and Satinder Singh and Sebastian Flennerhag},
booktitle={Proceedings of the 11th International Conference on Learning Representations},
year={2023},
address={Kigali, Rwanda},
}

@inproceedings{optformer,
author = {Chen, Yutian and Song, Xingyou and Lee, Chansoo and Wang, Zi and Zhang, Qiuyi and Dohan, David and Kawakami, Kazuya and Kochanski, Greg and Doucet, Arnaud and Ranzato, Marc'aurelio and Perel, Sagi and de Freitas, Nando},
title = {Towards learning universal hyperparameter optimizers with {T}ransformers},
booktitle = {Advances in Neural Information Processing Systems 36},
pages={32053--32068},
year={2022},
address = {New Orleans, LA}
}

@inproceedings{song2024ribbo,
  title={Reinforced In-Context Black-Box Optimization},
  author={Song, Lei and Gao, Chenxiao and Xue, Ke and Wu, Chenyang and Li, Dong and Hao, Jianye and Zhang, Zongzhang and Qian, Chao},
  booktitle={Proceedings of the 34th International Joint Conference on Artificial Intelligence},
  address={Montreal, Canada},
  year={2025},
pages={8939--8947},
}

@inproceedings{ela,
author = {Mersmann, Olaf and Bischl, Bernd and Trautmann, Heike and Preuss, Mike and Weihs, Claus and Rudolph, G\"{u}nter},
title = {Exploratory landscape analysis},
year = {2011},
booktitle = {Proceedings of the 13th Annual Conference on Genetic and Evolutionary Computation},
pages = {829--836},
address = {Dublin, Ireland},
}

@inproceedings{
neural-ela,
title={Neural Exploratory Landscape Analysis for Meta-Black-Box-Optimization},
author={Zeyuan Ma and Jiacheng Chen and Hongshu Guo and Yue-Jiao Gong},
booktitle={Proceedings of the 13th International Conference on Learning Representations},
year={2025},
address={Singapore}
}

@article{rl-das,
  author={Guo, Hongshu and Ma, Yining and Ma, Zeyuan and Chen, Jiacheng and Zhang, Xinglin and Cao, Zhiguang and Zhang, Jun and Gong, Yue-Jiao},
  journal={IEEE Transactions on Systems, Man, and Cybernetics: Systems}, 
  title={Deep Reinforcement Learning for Dynamic Algorithm Selection: A Proof-of-Principle Study on Differential Evolution}, 
  year={2024},
  volume={54},
  number={7},
  pages={4247-4259},
}

@inproceedings{as-2,
author = {Cenikj, Gjorgjina and Petelin, Ga\v{s}per and Eftimov, Tome},
title = {{TransOptAS}: Transformer-Based Algorithm Selection for Single-Objective Optimization},
year = {2024},
booktitle = {Proceedings of the Genetic and Evolutionary Computation Conference Companion},
pages = {403--406},
address = {Melbourne, Australia},
}

@article{ac-1,
author = {Hutter, Frank and Hoos, Holger H. and Leyton-Brown, Kevin and St\"{u}tzle, Thomas},
title = {{ParamILS}: An automatic algorithm configuration framework},
year = {2009},
journal = {Journal of Artificial Intelligence Research},
pages = {267--306},
}

@inproceedings{dac-1,
  author    = {A. Biedenkapp and H. F. Bozkurt and T. Eimer and F. Hutter and M. Lindauer},
  title     = {Dynamic Algorithm Configuration: Foundation of a New Meta-Algorithmic Framework},
  booktitle = {Proceedings of the 24th European Conference on Artificial Intelligence},
  year = {2020},
}

@inproceedings{madac,
author = {Xue, Ke and Xu, Jiacheng and Yuan, Lei and Li, Miqing and Qian, Chao and Zhang, Zongzhang and Yu, Yang},
title = {Multi-agent dynamic algorithm configuration},
year = {2022},
booktitle = {Advances in Neural Information Processing Systems 36},
address = {New Orleans, LA},
pages={20147--20161}
}

@inproceedings{sadn,
      title={Sequential Multi-Agent Dynamic Algorithm Configuration}, 
      author={Chen Lu and Ke Xue and Lei Yuan and Yao Wang and Yaoyuan Wang and Sheng Fu and Chao Qian},
      year={2025},
      booktitle={Advances in Neural Information Processing Systems 39},
address={Vancouver, Canada}
}

@InProceedings{PLR,
  title = 	 {Prioritized Level Replay},
  author =       {Jiang, Minqi and Grefenstette, Edward and Rockt{\"a}schel, Tim},
  booktitle = 	 {Proceedings of the 38th International Conference on Machine Learning},
  pages = 	 {4940--4950},
  year = 	 {2021},
address={Virtual}
}

@InProceedings{accel,
  title = 	 {Evolving Curricula with Regret-Based Environment Design},
  author =       {Parker-Holder, Jack and Jiang, Minqi and Dennis, Michael and Samvelyan, Mikayel and Foerster, Jakob and Grefenstette, Edward and Rockt{\"a}schel, Tim},
  booktitle = 	 {Proceedings of the 39th International Conference on Machine Learning},
  pages = 	 {17473--17498},
  year = 	 {2022},
}

@inproceedings{
in-the-zone,
title={In the {ZONE}: Measuring difficulty and progression in curriculum generation},
author={Rose E. Wang and Jesse Mu and Dilip Arumugam and Natasha Jaques and Noah Goodman},
booktitle={Deep Reinforcement Learning Workshop at NeurIPS 2022},
year={2022},
}

@article{pflacco,
    author = {Prager, Raphael Patrick and Trautmann, Heike},
    title = {{Pflacco}: Feature-Based Landscape Analysis of Continuous and Constrained Optimization Problems in {P}ython},
    journal = {Evolutionary Computation},
    pages = {1-25},
    year = {2023},
}

@inproceedings{robust-plr,
author = {Jiang, Minqi and Dennis, Michael and Parker-Holder, Jack and Foerster, Jakob and Grefenstette, Edward and Rockt\"{a}schel, Tim},
title = {Replay-guided adversarial environment design},
year = {2021},
booktitle = {Advances in Neural Information Processing Systems 35},
pages={1884--1897},
address={Virtual}
}

@inproceedings{DSAGE,
author = {Bhatt, Varun and Tjanaka, Bryon and Fontaine, Matthew C. and Nikolaidis, Stefanos},
title = {Deep surrogate assisted generation of environments},
year = {2022},
booktitle = {Advances in Neural Information Processing Systems 36},
address = {New Orleans, LA},
pages = {37762--37777}
}

@inproceedings{bengio2009curriculum,
author = {Bengio, Yoshua and Louradour, J\'{e}r\^{o}me and Collobert, Ronan and Weston, Jason},
title = {Curriculum learning},
year = {2009},
booktitle = {Proceedings of the International Conference on Machine Learning},
pages = {41--48},
address = {Montreal, Canada},
}

@article{schaul2015prioritized,
  title={Prioritized Experience Replay},
  author={Schaul, Tom and Quan, John and Antonoglou, Ioannis and Silver, David},
  journal={arXiv:1511.05952},
  year={2015}
}

@inproceedings{fitness-distance-corr,
author = {Jones, Terry and Forrest, Stephanie},
title = {Fitness Distance Correlation as a Measure of Problem Difficulty for Genetic Algorithms},
year = {1995},
address = {San Francisco, CA},
booktitle = {Proceedings of the 6th International Conference on Genetic Algorithms},
pages = {184--192},
}

@ARTICLE{nfl1,
  author={Wolpert, D.H. and Macready, W.G.},
  journal={IEEE Transactions on Evolutionary Computation}, 
  title={No free lunch theorems for optimization}, 
  year={1997},
  volume={1},
  number={1},
  pages={67-82},}

@inproceedings{PAIRED,
author = {Dennis, Michael and Jaques, Natasha and Vinitsky, Eugene and Bayen, Alexandre and Russell, Stuart and Critch, Andrew and Levine, Sergey},
title = {Emergent complexity and zero-shot transfer via unsupervised environment design},
year = {2020},
booktitle = {Advances in Neural Information Processing Systems 34},
address = {Vancouver, Canada},
}

@article{cmaes,
  title={Reducing the time complexity of the derandomized evolution strategy with covariance matrix adaptation ({CMA-ES})},
  author={Hansen, Nikolaus and M{\"u}ller, Sibylle D and Koumoutsakos, Petros},
  journal={Evolutionary computation},
  volume={11},
  pages={1--18},
  year={2003},
}

@inproceedings{madde,
  title={Improving differential evolution through {B}ayesian hyperparameter optimization},
  author={Biswas, Subhodip and Saha, Debanjan and De, Shuvodeep and Cobb, Adam D and Das, Swagatam and Jalaian, Brian A},
  booktitle={Proceedings of the IEEE Congress of Evolutionary Computation},
  year={2021},
}

@book{ZPD,
  author    = {Vygotsky, L. S.},
  editor    = {Cole, Michael and John-Steiner, Vera and Scribner, Sylvia and Souberman, Ellen},
  title     = {Mind in Society: The Development of Higher Psychological Processes},
  publisher = {Harvard University Press},
  address   = {Cambridge, MA},
  year      = {1978},
}

@InProceedings{global-structure,
author="Lunacek, Monte
and Whitley, Darrell
and Sutton, Andrew",
title="The Impact of Global Structure on Search",
booktitle="Proceedings of the 10th International Conference on Parallel Problem Solving from Nature",
year="2008",
address="Berlin, Heidelberg",
pages="498--507",
}

@InProceedings{oss_vizier,
  title = 	 {Open Source {Vizier}: {D}istributed Infrastructure and {API} for Reliable and Flexible Blackbox Optimization},
  author =       {Song, Xingyou and Perel, Sagi and Lee, Chansoo and Kochanski, Greg and Golovin, Daniel},
  booktitle = 	 {Proceedings of the 1st International Conference on Automated Machine Learning (AutoML)},
  pages = 	 {8/1--17},
  year = 	 {2022},
}

@article{mabbob,
author = {Vermetten, Diederick and Ye, Furong and B\"{a}ck, Thomas and Doerr, Carola},
title = {{MA-BBOB}: A Problem Generator for Black-Box Optimization Using Affine Combinations and Shifts},
year = {2025},
issue_date = {March 2025},
volume = {5},
number = {1},
pages={1--19},
journal = {ACM Transactions on Evolutionary Learning Optimization},
}

@inproceedings{configx,
author = {Guo, Hongshu and Ma, Zeyuan and Chen, Jiacheng and Ma, Yining and Cao, Zhiguang and Zhang, Xinglin and Gong, Yue-Jiao},
title = {{ConfigX}: modular configuration for evolutionary algorithms via multitask reinforcement learning},
year = {2025},
  booktitle={Proceedings of the 39th AAAI Conference on Artificial Intelligence},
address={Philadelphia, PA},
pages={26982--26990}
}

@InProceedings{q-mamba,
  title = 	 {Meta-Black-Box-Optimization through Offline {Q}-function Learning},
  author =       {Ma, Zeyuan and Cao, Zhiguang and Jiang, Zhou and Guo, Hongshu and Gong, Yue-Jiao},
  booktitle = 	 {Proceedings of the 42nd International Conference on Machine Learning},
  pages = 	 {41807--41826},
  year = 	 {2025},
address={Vancouver, Canada}
}

@article{gen-env,
      title={{GenEnv}: Difficulty-Aligned Co-Evolution Between {LLM} Agents and Environment Simulators}, 
      author={Jiacheng Guo and Ling Yang and Peter Chen and Qixin Xiao and Yinjie Wang and Xinzhe Juan and Jiahao Qiu and Ke Shen and Mengdi Wang},
      year={2025},
      journal={arXiv:2512.19682},
}

@INPROCEEDINGS{DR,
  author={Tobin, Josh and Fong, Rachel and Ray, Alex and Schneider, Jonas and Zaremba, Wojciech and Abbeel, Pieter},
  booktitle={2017 IEEE/RSJ International Conference on Intelligent Robots and Systems}, 
  title={Domain randomization for transferring deep neural networks from simulation to the real world}, 
  year={2017},
  pages={23-30},}

@InProceedings{position-fundational,
  title = 	 {Position: {L}everage Foundational Models for Black-Box Optimization},
  author =       {Song, Xingyou and Tian, Yingtao and Lange, Robert Tjarko and Lee, Chansoo and Tang, Yujin and Chen, Yutian},
  booktitle = 	 {Proceedings of the 41st International Conference on Machine Learning (ICML)},
    address={Vienna, Austria},
  pages = 	 {46168--46180},
  year = 	 {2024},
}

@inproceedings{metabox-v1,
author={Ma, Zeyuan and Guo, Hongshu and Chen, Jiacheng and Li, Zhenrui and Peng, Guojun and Gong, Yue-Jiao and Ma, Yining and Cao, Zhiguang},
title={{MetaBox}: A Benchmark Platform for Meta-Black-Box Optimization with Reinforcement Learning},
booktitle = {Advances in Neural Information Processing Systems 37},
year={2023},
}

@inproceedings{metabox-v2,
title={{MetaBox-v2}: A Unified Benchmark Platform for Meta-Black-Box Optimization},
author={Zeyuan Ma and Yue-Jiao Gong and Hongshu Guo and Wenjie Qiu and Sijie Ma and Hongqiao Lian and Jiajun Zhan and Kaixu Chen and Chen Wang and Zhiyang Huang and Zechuan Huang and Guojun Peng and Ran Cheng and Yining Ma},
booktitle={Advances in Neural Information Processing Systems 39},
year={2025},
}

@inproceedings{shin2025offline,
  title={Offline Model-based Optimization for Real-World Molecular Discovery},
  author={Shin, Dong-Hee and Son, Young-Han and Lee, Hyun Jung and Lee, Deok-Joong and Kam, Tae-Eui},
  booktitle={Proceedings of the 42nd International Conference on Machine Learning},
address={Vancouver, Canada},
  pages={3485-3494},
  year={2025}
}

@inproceedings{
wiremask-bbo,
title={Macro Placement by Wire-Mask-Guided Black-Box Optimization},
author={Yunqi Shi and Ke Xue and Lei Song and Chao Qian},
booktitle={Advances in Neural Information Processing Systems 37},
pages={6825--6843},
year={2023},
address={New Orleans, LA}
}

@article{bboplace-bench,
      title={{BBOPlace-Bench}: Benchmarking Black-Box Optimization for Chip Placement}, 
      author={Ke Xue and Ruo-Tong Chen and Rong-Xi Tan and Xi Lin and Yunqi Shi and Siyuan Xu and Mingxuan Yuan and Chao Qian},
      year={2025},
      journal={arXiv:2510.23472},
}

@inproceedings{PFNs4BO,
author = {M\"{u}ller, Samuel and Feurer, Matthias and Hollmann, Noah and Hutter, Frank},
title = {{PFNs4BO}: {I}n-context learning for {B}ayesian optimization},
year = {2023},
booktitle = {Proceedings of the 40th International Conference on Machine Learning},
pages={25444--25470},
address = {Honolulu, HI},
}

@inproceedings{ExPT,
author = {Nguyen, Tung and Agrawal, Sudhanshu and Grover, Aditya},
title = {{ExPT}: {S}ynthetic pretraining for few-shot experimental design},
year = {2023},
address = {New Orleans, LA},
booktitle = {Advances in Neural Information Processing Systems 36},
pages={45856--45869}
}

@InProceedings{design-bench,
  title = 	 {{D}esign-{B}ench: {B}enchmarks for Data-Driven Offline Model-Based Optimization},
  author =       {Trabucco, Brandon and Geng, Xinyang and Kumar, Aviral and Levine, Sergey},
  booktitle = 	 {Proceedings of the 39th International Conference on Machine Learning},
  pages = 	 {21658--21676},
  year = 	 {2022},
  address = {Baltimore, MD},
}

@inproceedings{
meidani2024snip,
title={{SNIP}: Bridging Mathematical Symbolic and Numeric Realms with Unified Pre-training},
author={Kazem Meidani and Parshin Shojaee and Chandan K. Reddy and Amir Barati Farimani},
booktitle={Proceedings of the 12th International Conference on Learning Representations},
year={2024},
address={Vienna, Austria}
}

@inproceedings{
wang2025synthetic,
title={Synthetic Series-Symbol Data Generation for Time Series Foundation Models},
author={Wenxuan Wang and Kai Wu and Yujian Betterest Li and Dan Wang and Xiaoyu Zhang},
booktitle={Advances in Neural Information Processing Systems 39},
year={2025},
}

@article{PPO,
  title={Proximal policy optimization algorithms},
  author={Schulman, John and Wolski, Filip and Dhariwal, Prafulla and Radford, Alec and Klimov, Oleg},
  journal={arXiv:1707.06347},
  year={2017}
}

@inproceedings{transformer,
  title={Attention is all you need},
  author={Vaswani, A and Shazeer, N and Parmar, N and Uszkoreit, J and Jones, L and Gomez, A and Kaiser, L and Polosukhin, I},
  booktitle={Advances in Neural Information Processing Systems 31},
  pages={6000--6010},
  address={Long Beach, CA},
  year={2017}
}

@book{gpml,
    title={{G}aussian {P}rocesses for {M}achine {L}earning},
    author ={Rasmussen, C. E. and Williams, C. K. I.},
    publisher ={The MIT Press},
    address={Cambridge, MA},
    year={2006}
}

@inproceedings{adam,
  title={Adam: {A} method for stochastic optimization},
  author={Diederik P. Kingma and Jimmy Ba},
  booktitle={Proceedings of the 3rd International Conference on Learning Representations},
  address = {San Diego, CA},
  year={2015}
}

@article{gym,
  title={Open{AI} {G}ym},
  author={Greg Brockman and Vicki Cheung and Ludwig Pettersson and Jonas Schneider and John Schulman and Jie Tang and Wojciech Zaremba},
  journal={arXiv:1606.01540},
  year={2016},
}

@InProceedings{robel,
  title = 	 {{ROBEL}: {R}obotics Benchmarks for Learning with Low-Cost Robots},
  author =       {Ahn, Michael and Zhu, Henry and Hartikainen, Kristian and Ponte, Hugo and Gupta, Abhishek and Levine, Sergey and Kumar, Vikash},
  booktitle = 	 {Proceedings of the 4th Conference on Robot Learning},
  pages = 	 {1300--1313},
  year = 	 {2020},
  address = {Virtual}, 
}

@article{tfbind,
  title={Survey of variation in human transcription factors reveals prevalent {DNA} binding changes},
  author={Luis A. Barrera and Anastasia Vedenko and Jesse V. Kurland and Julia M. Rogers and Stephen S. Gisselbrecht and Elizabeth J. Rossin and others},
  journal={Science},
  volume={351},
  number={6280},
  pages={1450--1454},
  year={2016},
}

@inproceedings{ioh,
author = {Doerr, Carola and Ye, Furong and Horesh, Naama and Wang, Hao and Shir, Ofer M. and B\"{a}ck, Thomas},
title = {Benchmarking discrete optimization heuristics with {IOHprofiler}},
year = {2019},
booktitle = {Proceedings of the Genetic and Evolutionary Computation Conference Companion},
pages = {1798--1806},
address = {Prague, Czech Republic},
}

@article{lde,
  author={Sun, Jianyong and Liu, Xin and B{\"a}ck, Thomas and Xu, Zongben},
  journal={IEEE Transactions on Evolutionary Computation},
  title={Learning Adaptive Differential Evolution Algorithm From Optimization Experiences by Policy Gradient},
  year={2021},
  volume={25},
  number={4},
  pages={666--680},
}

@inproceedings{b2opt,
  title={{B2Opt}: Learning to Optimize Black-box Optimization with Little Budget},
  author={Xiaobin Li and Kai Wu and Xiaoyu Zhang and Handing Wang},
  booktitle={Proceedings of the 39th AAAI Conference on Artificial Intelligence},
  pages={18502--18510},
  year={2025},
  address={Philadelphia, PA},
}

@inproceedings{glhf,
  title={Pretrained Optimization Model for Zero-Shot Black Box Optimization},
  author={Xiaobin Li and Kai Wu and Yujian Betterrest Li and Xiaoyu Zhang and Handing Wang and Jing Liu},
  booktitle={Advances in Neural Information Processing Systems},
  volume={37},
  pages={14283--14324},
  year={2024},
  address={Vancouver, Canada},
}

@inproceedings{samplr,
  title={Grounding Aleatoric Uncertainty for Unsupervised Environment Design},
  author={Jiang, Minqi and Dennis, Michael and Parker-Holder, Jack and Lupu, Andrei and K{\"u}ttler, Heinrich and Grefenstette, Edward and Rockt{\"a}schel, Tim and Foerster, Jakob},
  booktitle={Advances in Neural Information Processing Systems 35},
  pages={32868--32881},
  year={2022},
  address={New Orleans, LA},
}

@inproceedings{no-regrets-ued,
  title={No Regrets: Investigating and Improving Regret Approximations for Curriculum Discovery},
  author={Rutherford, Alexander and Beukman, Michael and Willi, Timon and Lacerda, Bruno and Hawes, Nick and Foerster, Jakob},
  booktitle={Advances in Neural Information Processing Systems 37},
  pages={16071--16101},
  year={2024},
  address={Vancouver, Canada},
}
\bibliographystyle{iclr2027_conference}

\newpage
\appendix
\section{Additional Background}
\label{appendix:background}
\subsection{Further Developments in Automated Curriculum Learning}
\label{appendix:acl-background}

Several works further extend UED and prioritized replay to more complex settings.
PLR$^{\perp}$~\citep{robust-plr} extends PLR to stochastic environments by reducing noise from epistemic uncertainty.
ACCEL~\citep{accel} further uses evolutionary strategies to progressively mutate and generate more complex environments from the PLR buffer~\citep{DSAGE}.
More recently, these UED principles have also been applied to more complex domains, such as generating synthetic environments for LLM agents~\citep{gen-env}.

\section{A Brief Introduction of the Employed Method Backbone}

\subsection{SYMBOL}
\label{appendix:detail-symbol}
SYMBOL~\citep{chen2024symbol} is a MetaBBO framework that learns optimizers by symbolically generating update rules. Instead of using a fixed hand-crafted algorithm, SYMBOL trains a symbolic equation generator, parameterized by RL policy, to output an explicit update expression built from a predefined set of basic operators and optimization-related operands. The generated symbolic expression is then executed as the optimizer's update rule to propose new candidate solutions during search, which yields an adaptive yet interpretable optimization procedure.

Practically, the equation generator is meta-trained over the CEC function class~\citep{cec2021}, which is augmented with random shift and rotation.
The policy is trained via Proximal Policy Optimization~\citep{PPO}.
The reward consists of two parts: A base reward indicating the optimizer's absolute performance and a guided reward forcing the optimize the mimic the behavior of a teacher algorithm, typically instantiated by MadDE~\citep{madde}. 

\subsection{ExPT}
\label{appendix:detail-expt}
ExPT~\citep{ExPT} is a pretraining-adaptation framework for few-shot offline BBO, i.e., optimizing the black-box function $f$ using only a few-shot dataset $\{\x_i,f(\x_i)\}_{i=1}^N$. 
ExPT pretrains a transformer model~\citep{transformer} with a large amount of synthetic data, so that it can adapt to a new objective using only a few labeled observed $(x,y)$ pairs. 
ExPT uses \emph{inverse modeling}: given a small set of context points $(\x_{1:m},y_{1:m})$ and a set of desired target outputs $y_{m+1:N}$, the model approximates $p(\x_{m+1:N}\mid\x_{1:m},y_{1:m},y_{m+1:N})$, and samples candidate $\x$ during inference.

To obtain substantial pretraining data, ExPT samples synthetic functions from a Gaussian process prior with an RBF kernel~\citep{gpml},
$\tilde{f}\sim \mathcal{GP}(0, K)$ where $K(x,x')=\sigma^2\exp\!\left(-\frac{\|x-x'\|^2}{2\ell^2}\right)$.
Here $\sigma$ (variance) controls the scale of function values and $\ell$ (length scale) controls smoothness.
Specifically, ExPT randomizes $\sigma$ and $\ell$ to increase function diversity. 
For each sampled function, it evaluates $\tilde{f}$ on a set of unlabeled inputs, then splits the resulting input-output pairs into a small context set and a target set to form training examples.

\section{Implementation Details}
\label{appendix:implementation-details}

\subsection{Detailed Setting and Hyperparameters for HATS}
To quantify the utility performance gap in Equation~\eqref{eq:gap}, following SYMBOL~\citep{chen2024symbol}, we select MadDE~\citep{madde} as the reference optimizer to evaluate the generated instances; notably, in HATS it only provides a reference metric for scoring tasks rather than participating in policy training.
Other hyperparameters for HATS are listed in Table~\ref{tab:hpm}.

\begin{table}[h]
    \centering
    \caption{Hyperparameter setting for HATS.}
    \begin{tabular}{c|c}
        \toprule
        Parameter Name & Values \\
        \midrule
        Utility range $[u_{\min},u_{\max}]$ & [-0.5, 4.0] \\
        Minimum class probability $\epsilon$ (CEC / BBOB) & 0.05 / 0.02 \\
        Decay rate $\gamma$ for bandit logits & 0.9 \\
        Update rate $\eta$ for bandit logits & 0.1 \\
        Clipping bound $L$ for bandit logits & 1.5 \\
        Balancing coefficient $\alpha$ for staleness in replay buffer & 0.1 \\
        \midrule 
        Optimizer & Adam~\citep{adam} \\
        Learning rate & $1\times 10^{-3}$ \\
        \bottomrule
    \end{tabular}
    \label{tab:hpm}
\end{table}

\subsection{Implementation Details of Benchmarks, Parameter Generator, and Compared Baselines}
For BBOB~\citep{bbob}, we use the $24$ function classes of $10$D problems from MetaBox~\citep{metabox-v2}~\footnote{\url{https://github.com/MetaEvo/MetaBox/tree/v2.0.0/src/environment/problem/SOO/COCO_BBOB}}, where the parameter generator in Algorithm~\ref{alg:robust-plr} (Line 5) generates new instances by randomly sampling shift vectors and rotation matrices $\bds{\theta} \in \Theta$ following MetaBox~\footnote{\url{https://github.com/MetaEvo/MetaBox/blob/v2.0.0/src/environment/problem/SOO/COCO_BBOB/bbob_numpy.py}}.
For CEC~\citep{cec2021}, we directly use the $10$ function classes of $10$D problems provided by SYMBOL~\citep{chen2024symbol}~\footnote{\url{https://github.com/MetaEvo/Symbol/blob/main/dataset/cec_dataset.py}} and sample shift and rotation parameters following SYMBOL's instance generation routine~\footnote{\url{https://github.com/MetaEvo/Symbol/blob/main/dataset/generate_dataset.py}}.
For evaluation on BBOB and CEC, we use held-out $10$D instances generated with unseen shifts and rotations under fixed seeds. For realistic tasks in MetaBox, we use the \texttt{difficulty=all} split with their problem-specific dimensions ($12$D for Protein Docking~\citep{protein} and $30$D for UAV~\citep{uav}).

For the compatible PLR-style baselines in Table~\ref{tab:main-results}, PLR$^{\perp}$-L1 and HATS-L1 compute the L1 value loss $|\hat{R}_t - V(s_t)|$~\citep{PLR} using SYMBOL's native $n$-step return targets while keeping its PPO training objective unchanged.

For MA-BBOB~\citep{mabbob}, we use the implementation by IOHProfiler~\citep{ioh} by randomly selecting a set of 24 function instances and train the agent using this set.

For Diverse-BBO, we directly use the \texttt{pickle} file provided by~\citep{diverse-bbo}~\footnote{\url{https://github.com/MetaEvo/Diverse-BBO/blob/main/save_gp_functions/all_256_programs.pickle}}. However, we find that the provided function instances are prone to outliers due to the symbolic function property, which may cause damage the training process. Therefore, we take a dynamic strategy that once we meet an outlier that cause numerical overflow, we filter out the current instance and resume the checkpoint from last episode to enable training for Diverse-BBO.

\subsection{Implementation Details of HATS on ExPT}
\label{appendix:implement-expt}
At each pretraining iteration, we sample GP hyperparameters $\theta$ to generate synthetic functions and train ExPT on the
collected data from the functions. 
We define the utility of a sampled configuration $\theta$ by the training negative
log-likelihood: $U_{\text{ExPT}}(\theta) = -\log p_\phi(\x_{m+1:N}\mid\x_{1:m},y_{1:m},y_{m+1:N})$ where $p_\phi$ denotes the distribution parameterized by the model,
and use it as the feedback signal for utility-aware replay.
This follows the HATS principle that $U$ should reflect
the training value of an instance for the current model.

In our implementation, we apply HATS by scheduling the two core GP hyperparameters that govern instance generation and
difficulty: the lengthscale $\ell$ and the function scale $\sigma$.
Let
$
\theta=(\ell,\sigma),~
\ell \sim \mathcal{U}[1.0,10.0],~
\sigma \sim \mathcal{U}[5.0,10.0]$. These prior ranges are set following the ExPT. We perform utility-aware replay over
previously sampled $\theta$ for training.

Following ExPT, we evaluate the trained model on 4 tasks from Design-Bench~\citep{design-bench}, a prevalent benchmark suite for offline BBO. 
These tasks include 2 continuous tasks, Ant Morphology~\citep{gym} and D'Kitty Morphology~\citep{robel}, and 2 discrete tasks, TF-Bind-8 and TF-Bind-10~\citep{tfbind}. 
The objective of Design-Bench is to maximize the objective score, which is different to our main experiment in Table~\ref{tab:main-results}. 

\section{Additional Experimental Results}
\label{appendix:additional-exp}

\subsection{Controlled Continuation Across Utility Zones}
\label{appendix:zone-continuation}

Complementing the LOO correlation study in Section~\ref{sec:exp-results}, we further examine whether the proposed utility directly predicts subsequent learning gains through controlled continuation experiments. Specifically, we resume SYMBOL from checkpoints at epochs $25$, $50$, and $75$, and group candidate tasks according to their reference--agent log-gap utility $U$ in Equation~\eqref{eq:gap} into three zones: \emph{too-easy} ($U < -0.5$), \emph{in-zone} ($-0.5 \le U \le 4.0$), and \emph{too-hard} ($U > 4.0$). For each checkpoint, we randomly select $10$ tasks from each zone and continually train each model for $1{,}000$ episodes. Table~\ref{tab:zone-continuation} reports the relative performance gains over the starting policy on held-out test instances. Across all three checkpoints, \emph{in-zone} tasks consistently yield the largest improvements, whereas \emph{too-easy} tasks bring limited gains at later stages and \emph{too-hard} tasks even degrade mid- and late-stage policies. These results confirm that prioritizing tasks within a moderate reference--agent gap leads to more effective learning progress.

\begin{table}[!h]
\centering
\caption{Relative test performance gains ($\%$) over the starting SYMBOL policy after controlled continuation training ($1{,}000$ episodes on $10$ randomly sampled tasks per utility zone) from epochs $25$, $50$, and $75$.}
{\setlength{\tabcolsep}{5pt}%
\begin{tabular}{l|ccc}
\toprule
Starting Checkpoint & Too-easy ($U < -0.5$) & In-zone ($-0.5 \le U \le 4.0$) & Too-hard ($U > 4.0$) \\ \midrule
Epoch 25 & $+5.05\%$ & $\mathbf{+26.00\%}$ & $+15.33\%$ \\
Epoch 50 & $+16.51\%$ & $\mathbf{+18.26\%}$ & $-5.31\%$ \\
Epoch 75 & $+1.31\%$ & $\mathbf{+4.82\%}$ & $-7.29\%$ \\
\bottomrule
\end{tabular}}
\label{tab:zone-continuation}
\end{table}

\subsection{Sensitivity and Learning Dynamics with Respect to the Reference Optimizer}
\label{appendix:teacher-sensitivity}

\textbf{Sensitivity across Different Reference Optimizers.}
To examine how the choice of reference optimizer $\A_{\text{ref}}$ in Equation~\eqref{eq:gap} affects HATS, we apply three reference optimizers (CMA-ES~\citep{cmaes}, PSO, and MadDE~\citep{madde}) across four MetaBBO backbones (SYMBOL~\citep{chen2024symbol}, POM~\citep{glhf}, LDE~\citep{lde}, and B2Opt~\citep{b2opt}). As shown in Table~\ref{tab:teacher-sensitivity}, HATS is reasonably robust across different reference choices while showing a consistent trend across backbones: using MadDE as the reference generally achieves the best average rank. Because the reference shapes curriculum quality rather than imposing a performance ceiling, a stronger reference is not necessarily better; instead, one matching the current agent's capability range works best. An overly strong reference (e.g., CMA-ES on synthetic functions) may make too many tasks appear overly hard, whereas a moderately strong reference such as MadDE remains closer to the agent's learnable zone and yields a more informative curriculum.

\begin{table*}[!h]
\centering
\caption{Sensitivity of HATS to different reference optimizers (CMA-ES, PSO, and MadDE) across MetaBBO backbones. Lower normalized scores and ranks indicate better performance; B2Opt is omitted from real-world tasks due to its dimension-dependent model trained under $10$D.}
\resizebox{\linewidth}{!}{%
{\setlength{\aboverulesep}{0pt}\setlength{\belowrulesep}{0pt}%
\renewcommand{\arraystretch}{1.15}%
\begin{tabular}{ll|cccc|cc}
\toprule
Backbone & Reference & CEC & BBOB & Protein Docking & UAV & Avg. Score & Avg. Rank \\ \midrule
\rowcolor{gray!15}\multicolumn{8}{c}{\textbf{Reference Baselines}} \\ \midrule
-- & CMA-ES & 0.026 $\pm$ 0.043 & 0.014 $\pm$ 0.037 & 0.887 $\pm$ 0.068 & 0.904 $\pm$ 0.275 & 0.458 & 3.28 \\
-- & PSO & 0.079 $\pm$ 0.116 & 0.062 $\pm$ 0.065 & 0.879 $\pm$ 0.069 & 0.908 $\pm$ 0.258 & 0.482 & 6.10 \\
-- & MadDE & 0.109 $\pm$ 0.138 & 0.104 $\pm$ 0.095 & 0.850 $\pm$ 0.085 & 0.908 $\pm$ 0.246 & 0.493 & 7.69 \\ \midrule
\rowcolor{gray!15}\multicolumn{8}{c}{\textbf{HATS with Different References}} \\ \midrule
\multirow{3}{*}{SYMBOL} & CMA-ES & 0.109 $\pm$ 0.120 & 0.114 $\pm$ 0.104 & 0.894 $\pm$ 0.063 & 0.930 $\pm$ 0.205 & 0.511 & 8.57 \\
 & PSO & 0.184 $\pm$ 0.177 & 0.185 $\pm$ 0.122 & 0.903 $\pm$ 0.058 & 0.939 $\pm$ 0.176 & 0.553 & 10.40 \\
 & MadDE & 0.095 $\pm$ 0.122 & 0.127 $\pm$ 0.145 & 0.855 $\pm$ 0.085 & 0.905 $\pm$ 0.268 & \textbf{0.495} & \textbf{7.57} \\ \midrule
\multirow{3}{*}{POM} & CMA-ES & 0.373 $\pm$ 0.243 & 0.401 $\pm$ 0.214 & 0.027 $\pm$ 0.053 & 0.077 $\pm$ 0.149 & \textbf{0.220} & 10.43 \\
 & PSO & 0.370 $\pm$ 0.234 & 0.400 $\pm$ 0.216 & 0.029 $\pm$ 0.061 & 0.117 $\pm$ 0.228 & 0.229 & 10.43 \\
 & MadDE & 0.375 $\pm$ 0.243 & 0.407 $\pm$ 0.215 & 0.005 $\pm$ 0.015 & 0.108 $\pm$ 0.198 & 0.224 & \textbf{10.41} \\ \midrule
\multirow{3}{*}{LDE} & CMA-ES & 0.024 $\pm$ 0.051 & 0.016 $\pm$ 0.030 & 0.911 $\pm$ 0.055 & 0.897 $\pm$ 0.277 & 0.462 & 5.15 \\
 & PSO & 0.016 $\pm$ 0.033 & 0.013 $\pm$ 0.027 & 0.905 $\pm$ 0.066 & 0.900 $\pm$ 0.270 & \textbf{0.458} & 4.62 \\
 & MadDE & 0.023 $\pm$ 0.047 & 0.015 $\pm$ 0.033 & 0.901 $\pm$ 0.061 & 0.897 $\pm$ 0.276 & 0.459 & \textbf{4.04} \\ \midrule
\multirow{3}{*}{B2Opt} & CMA-ES & 0.116 $\pm$ 0.143 & 0.106 $\pm$ 0.096 & / & / & 0.111 & 8.57 \\
 & PSO & 0.121 $\pm$ 0.154 & 0.109 $\pm$ 0.095 & / & / & 0.115 & 8.90 \\
 & MadDE & 0.093 $\pm$ 0.140 & 0.074 $\pm$ 0.076 & / & / & \textbf{0.083} & \textbf{6.59} \\
\bottomrule
\end{tabular}}}
\label{tab:teacher-sensitivity}
\end{table*}

\begin{figure}[!h]
    \centering
    \includegraphics[width=0.325\linewidth]{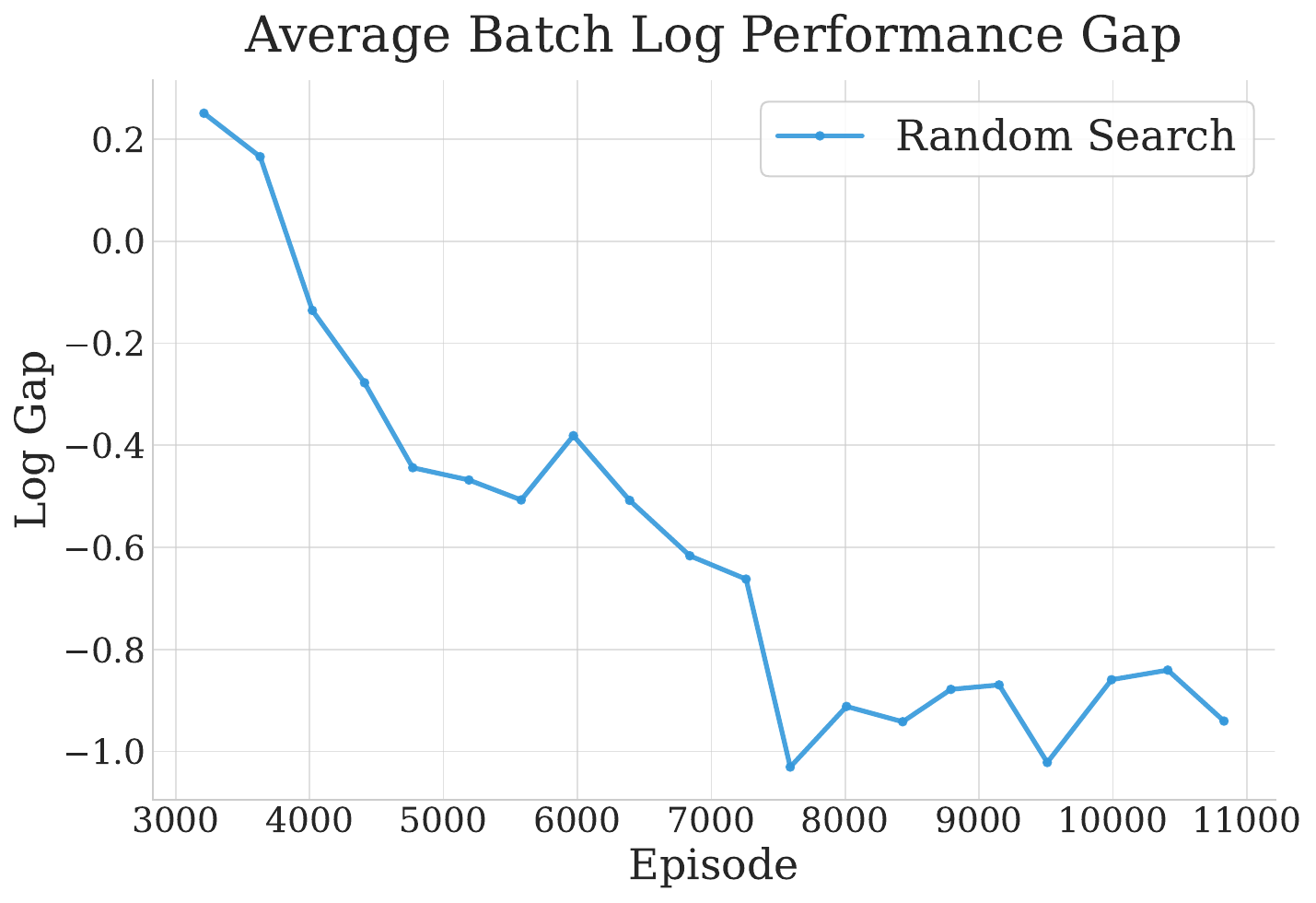}\hfill
    \includegraphics[width=0.325\linewidth]{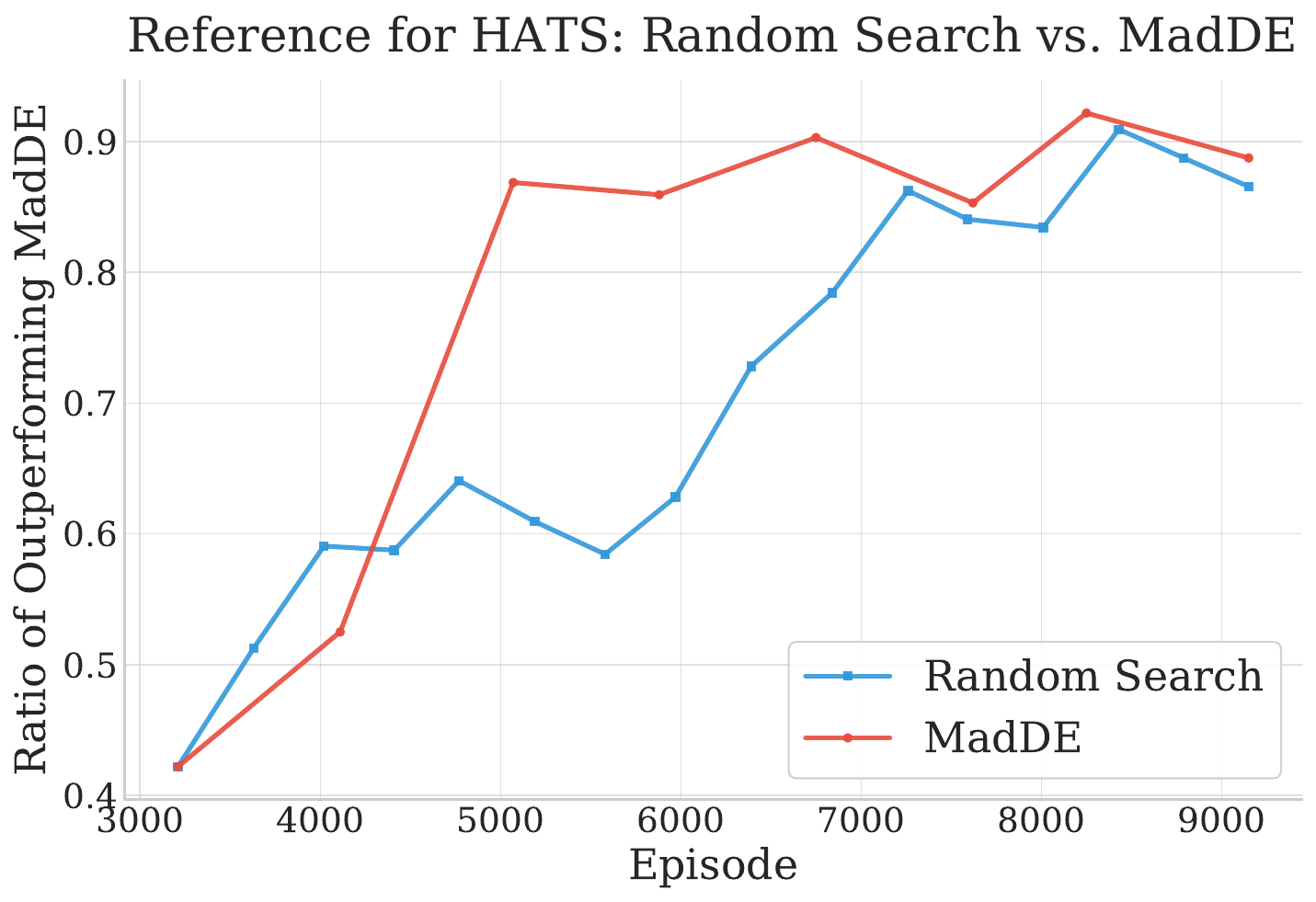}\hfill
    \includegraphics[width=0.325\linewidth]{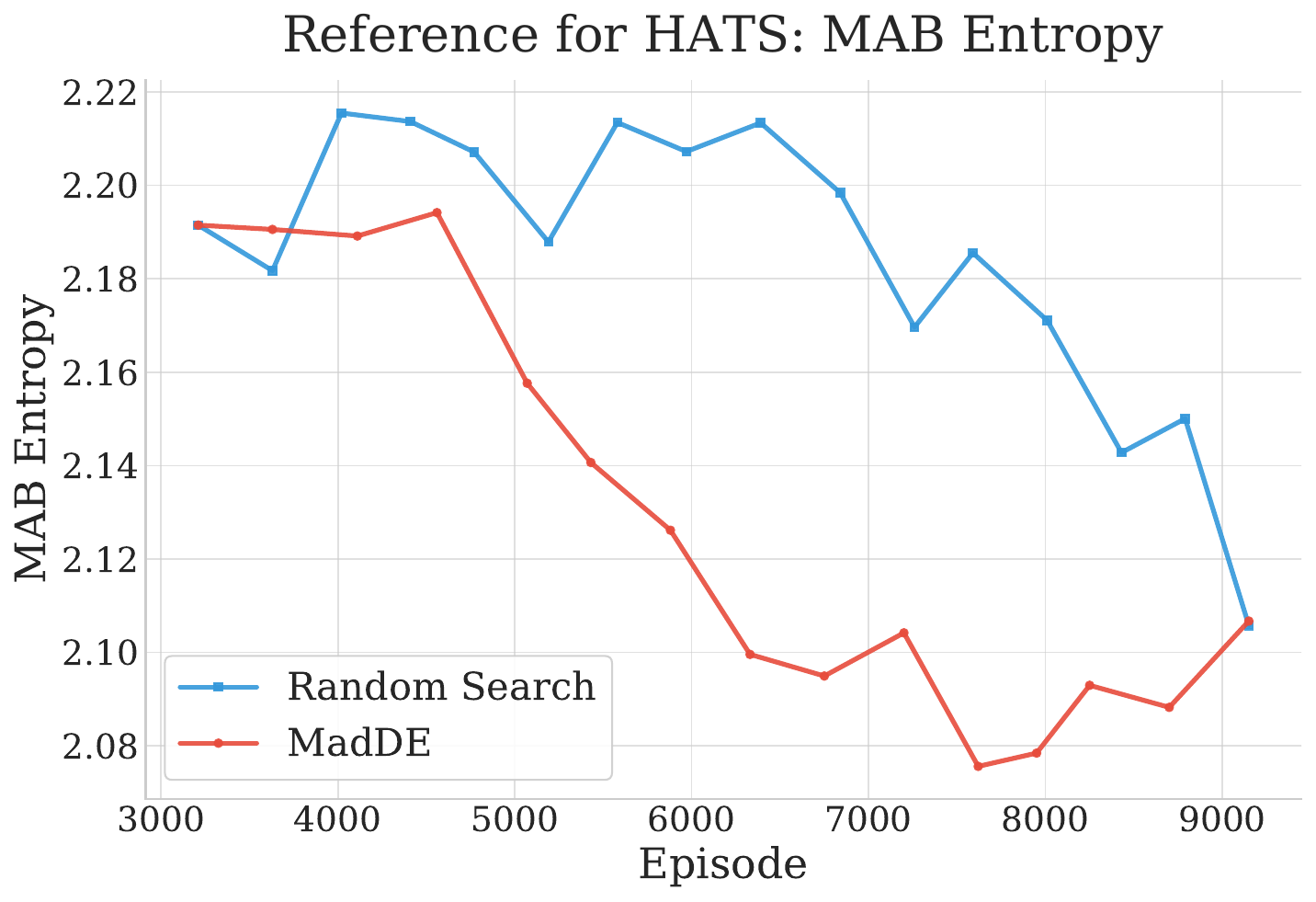}
    \caption{Training dynamics when replacing the MadDE reference with Random Search at epoch 5: (left) batch-average log performance gap, (middle) outperform ratio on held-out CEC test instances, and (right) MAB class-distribution entropy.}
    \label{fig:teacher-random-switch}
\end{figure}

\textbf{Continued Improvement After Surpassing a Weak Reference.}
We also examine whether reference-based selection restricts further learning once the agent surpasses the reference. In HATS, the reference optimizer only provides a reference metric to score and prioritize instances that the agent has not fully mastered, rather than directly training the policy via imitation or distillation; policy training is determined entirely by the MetaBBO backbone itself. To verify this, we replace the reference optimizer with \emph{Random Search} at the epoch-5 checkpoint and continue training for $20$ epochs. As shown in Figure~\ref{fig:teacher-random-switch}, although the batch-average log-gap becomes negative (left), the agent's outperform ratio on held-out CEC test instances continues to improve steadily and eventually reaches a level similar to the MadDE-referenced run (albeit with slower mid-stage convergence) since both train the same backbone (middle), while the MAB entropy does not collapse (right). These results confirm that the learned agent is not capped by a weak reference and can continue improving after surpassing it. We leave adaptive reference selection to future work.

\subsection{Analysis of Robust MAB}
\label{appendix:robust-mab}

\begin{figure}[!h]
    \centering
    \includegraphics[width=0.49\linewidth]{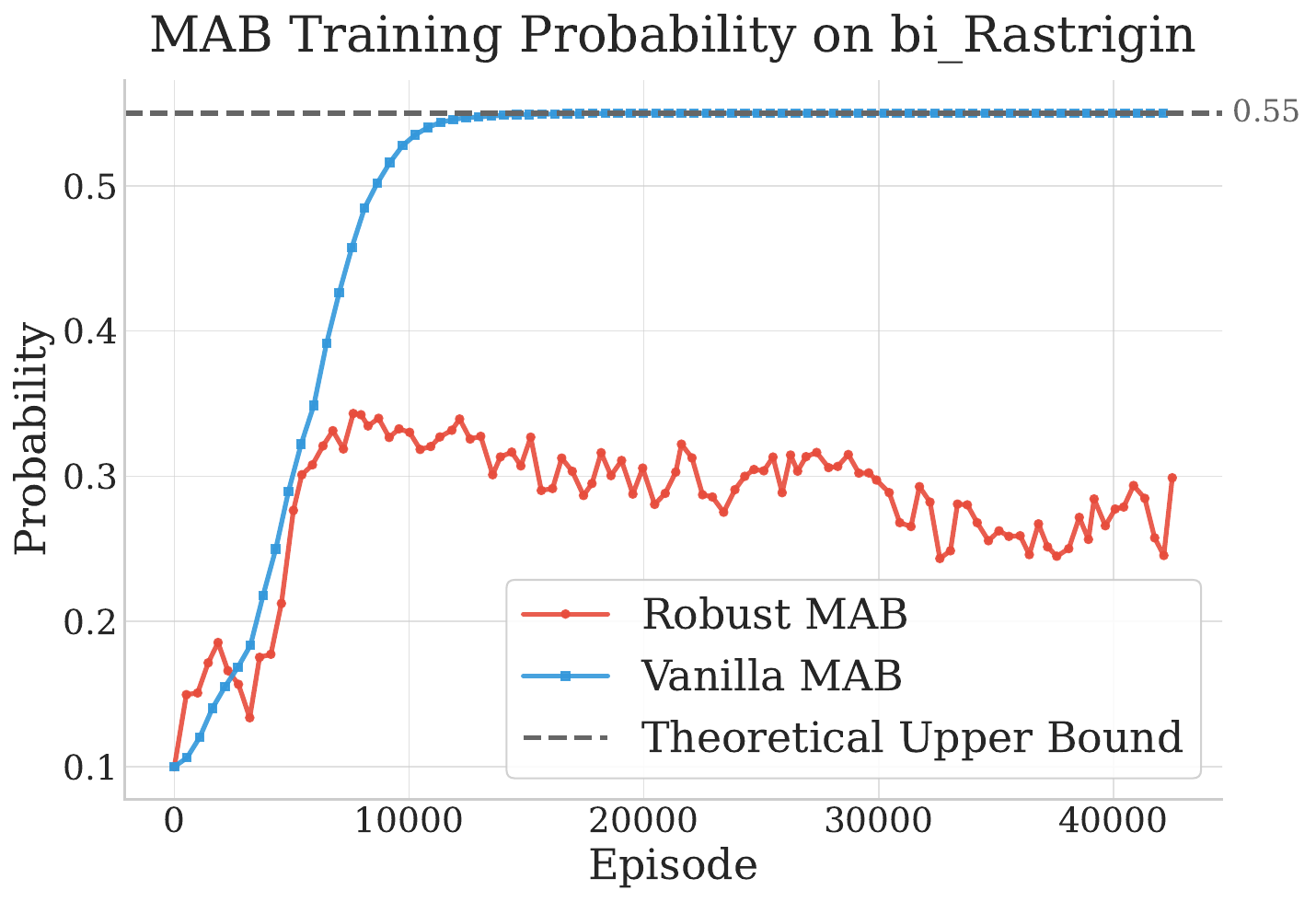}
    \includegraphics[width=0.49\linewidth]{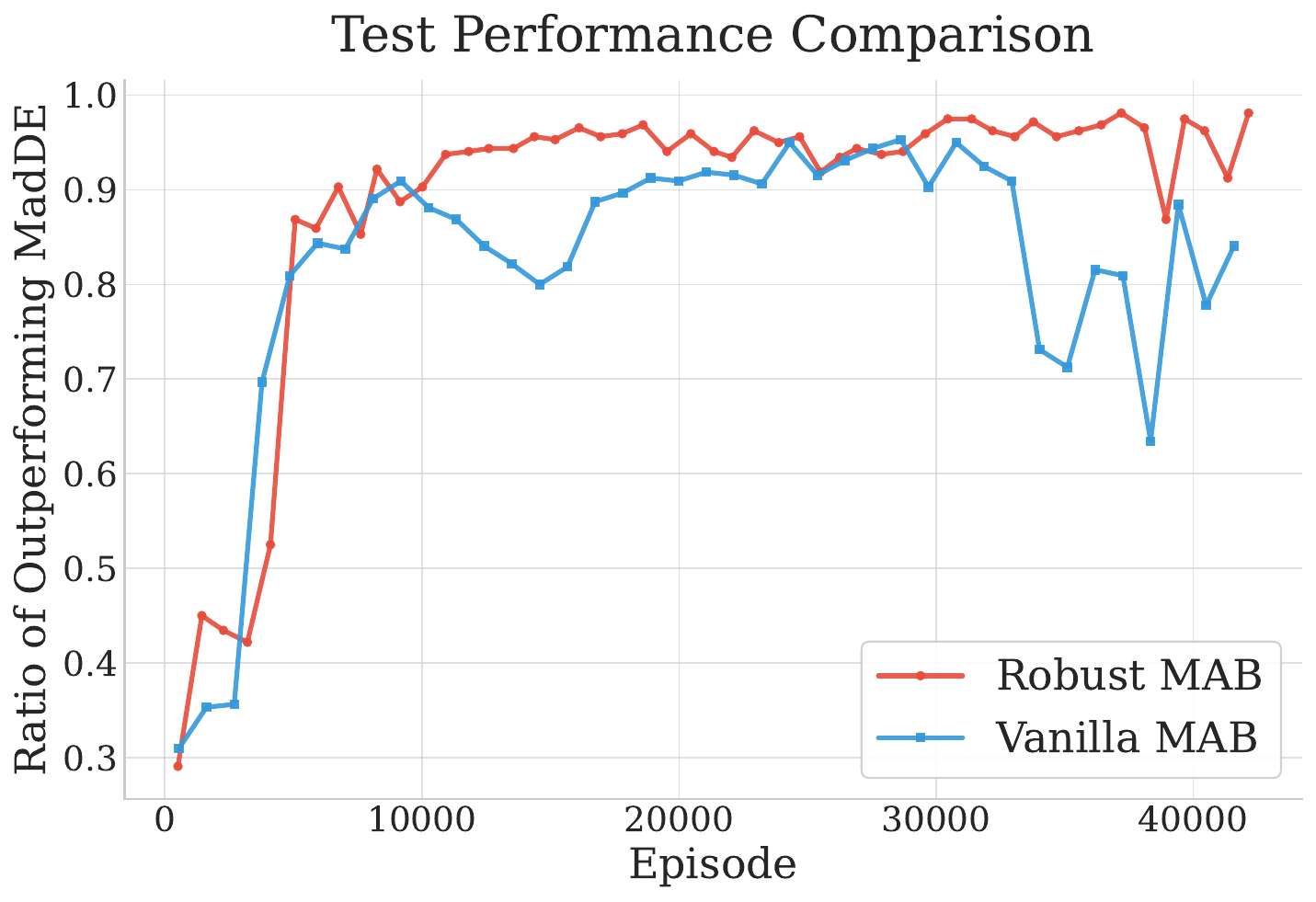}
    \caption{Learning dynamics of Robust MAB and Vanilla MAB. The left sub-figure plots MAB probability of choosing the \texttt{bi\_Rastrigin} function, while the right one displays the test performance throughout training.}
    \label{fig:mab-robust}
\end{figure}

We empirically find that deciding the function class solely based on a vanilla MAB can lead to probability collapse.
During the training of SYMBOL~\citep{chen2024symbol} on 10 CEC function classes~\citep{cec2021}, we find that the agent struggles on the \texttt{bi\_Rastrigin} function.
As training progresses, the probability of selecting \texttt{bi\_Rastrigin} converges to 0.55, the theoretical maximum probability of a function class, i.e., the MAB collapses to this arm.
This results in poor performance for the agent.
As shown in Figure~\ref{fig:mab-robust}, after 30000 episodes, the agent trained with vanilla MAB suffers from catastrophic forgetting.

In contrast, Robust MAB assigns a more balanced probability to the \texttt{bi\_Rastrigin} arm, showing smoother and more consistent improvement on the test set throughout training.

\subsection{Ablation Study: Does Each Component of HATS Contribute to the Final Performance?}
\label{sec:ablate}
\begin{figure}[!h]
    \centering
    \includegraphics[width=0.49\linewidth]{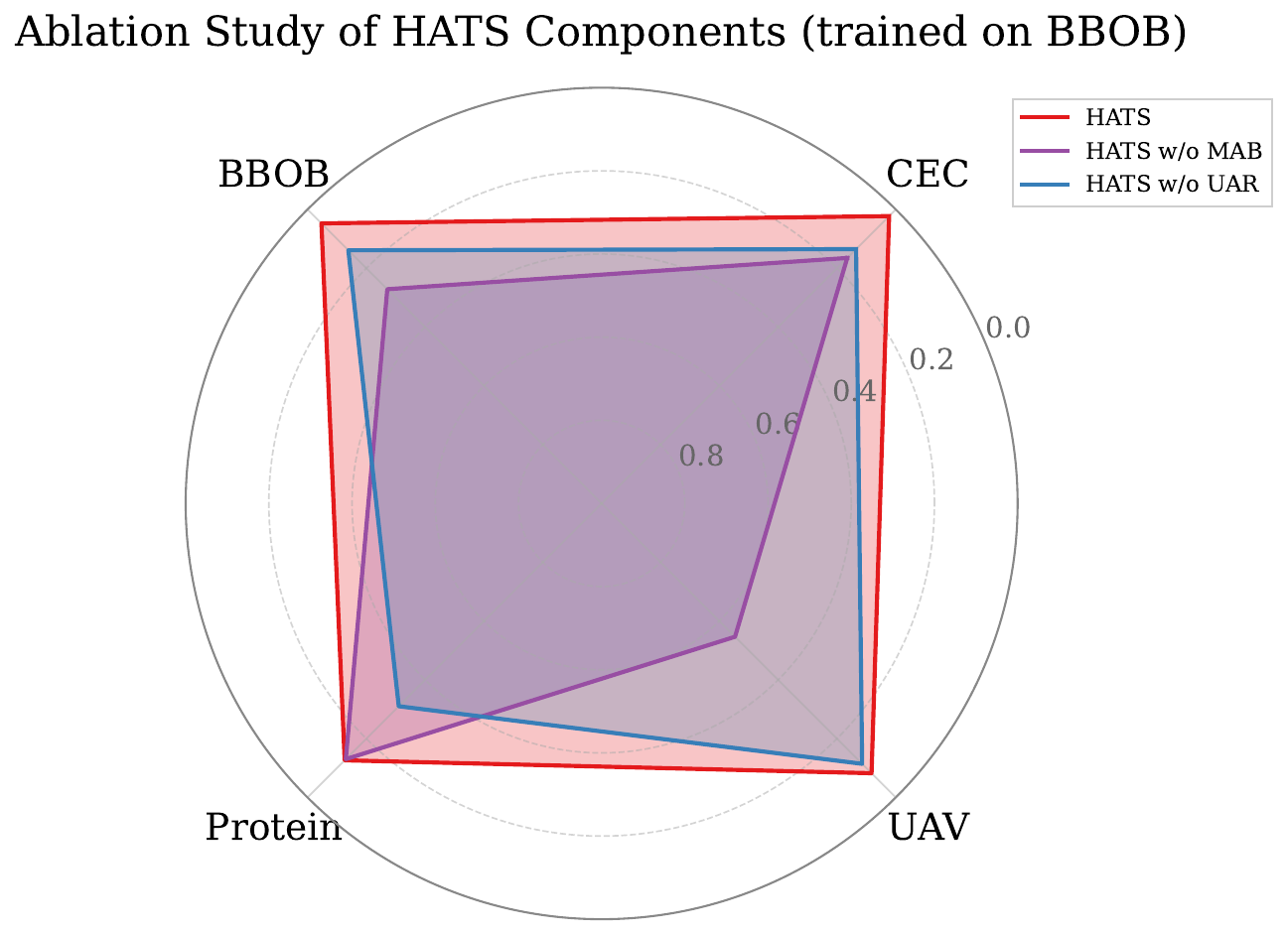}
    \includegraphics[width=0.49\linewidth]{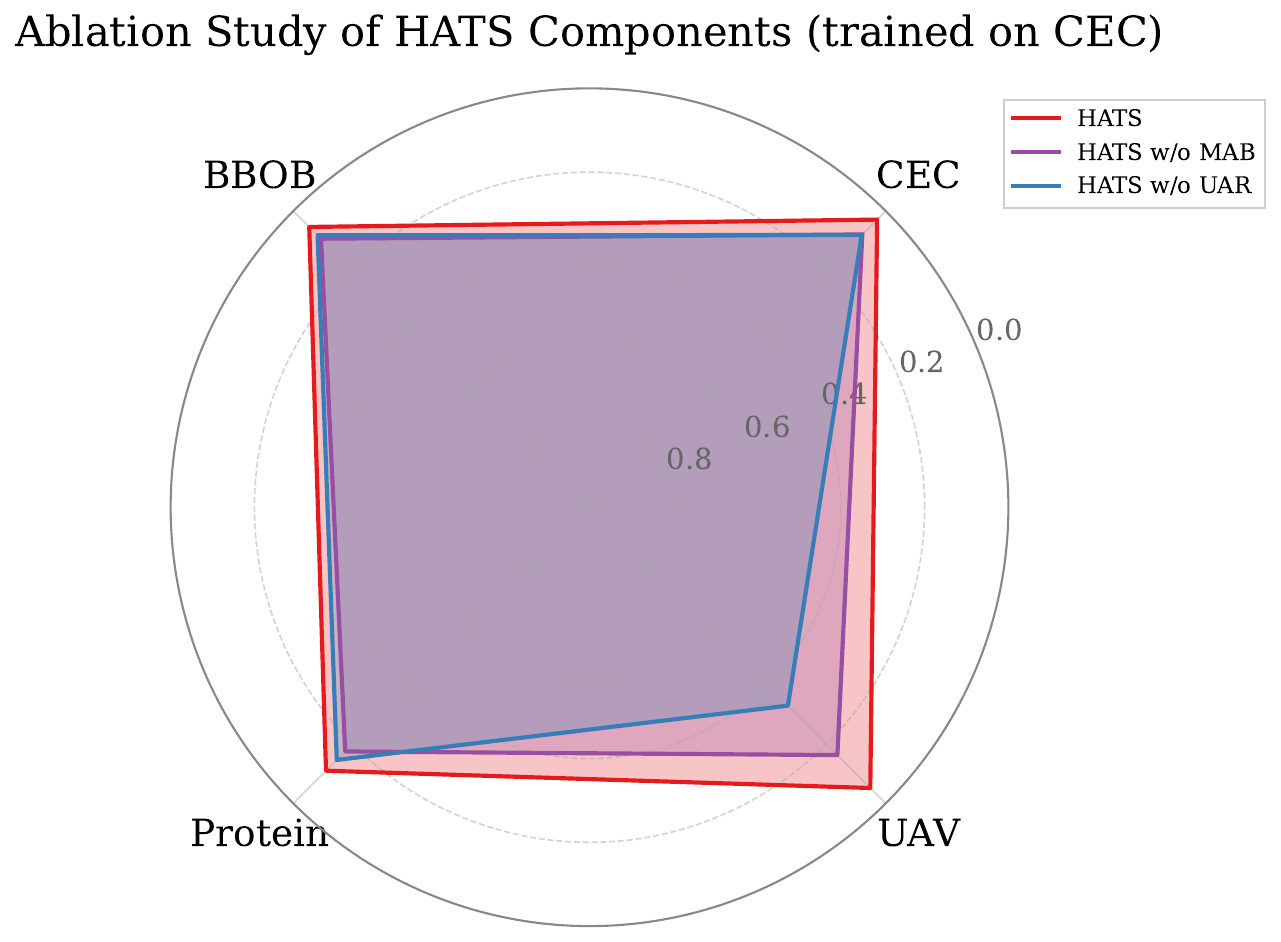}
    \caption{Ablation results on HATS components, MAB and UAR.}
    \label{fig:ablate}
\end{figure}

We ablate key components in HATS to isolate their contributions by:
(i) removing the multi-armed bandit for function class selection, (ii) removing utility-aware replay for instance parameters.
Experimental results in Figure~\ref{fig:ablate} clearly demonstrates the effectiveness of each component in HATS.

\subsection{Generalization to Few-Shot Offline BBO}
\label{appendix:expt-results}

\begin{figure}[!h]
    \centering
    \includegraphics[width=0.55\linewidth]{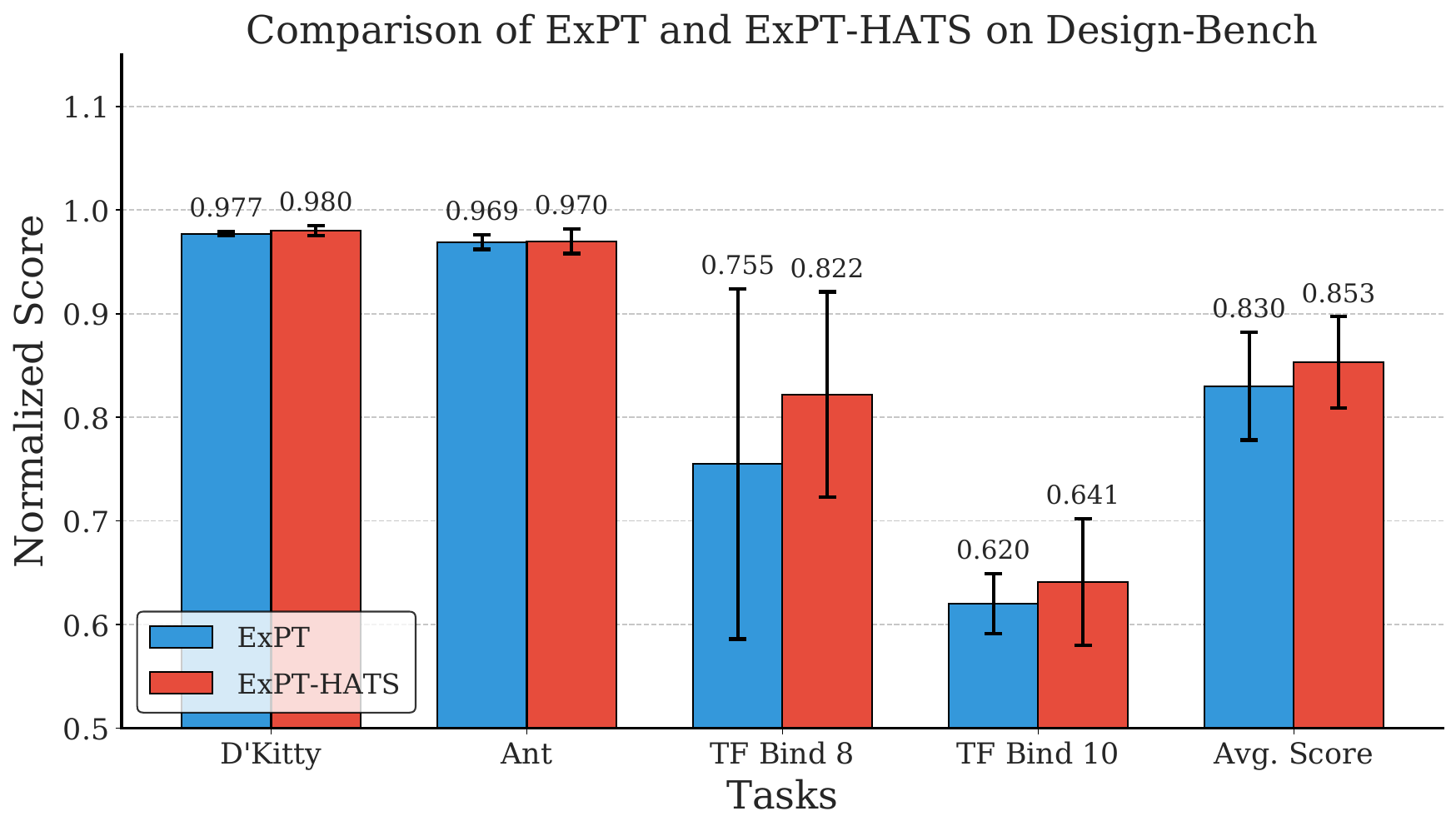}
    \caption{Normalized score (maximization) on Design-Bench~\citep{design-bench}, comparing ExPT and ExPT-HATS.}
    \label{fig:expt-results}
\end{figure}

We additionally examine the effectiveness of HATS on the few-shot data-driven offline BBO scenario.
We build upon ExPT~\citep{ExPT}, a framework that directly models the final solution conditioned on the few-shot data.
ExPT is pretrained under Gaussian processes with varying hyperparameters.
We apply HATS to schedule these hyperparameters.
Details of ExPT and our HATS implementation can be found in Appendix~\ref{appendix:implement-expt}.
As shown in Figure~\ref{fig:expt-results}, combined with HATS, the performance of ExPT can be further enhanced, showing the versatility of HATS.

\subsection{Case Study: How does SYMBOL Combined with HATS Generalize Better than DR?}
\label{appendix:case-study}
\begin{figure}[!h]
    \centering
    \includegraphics[width=0.45\linewidth]{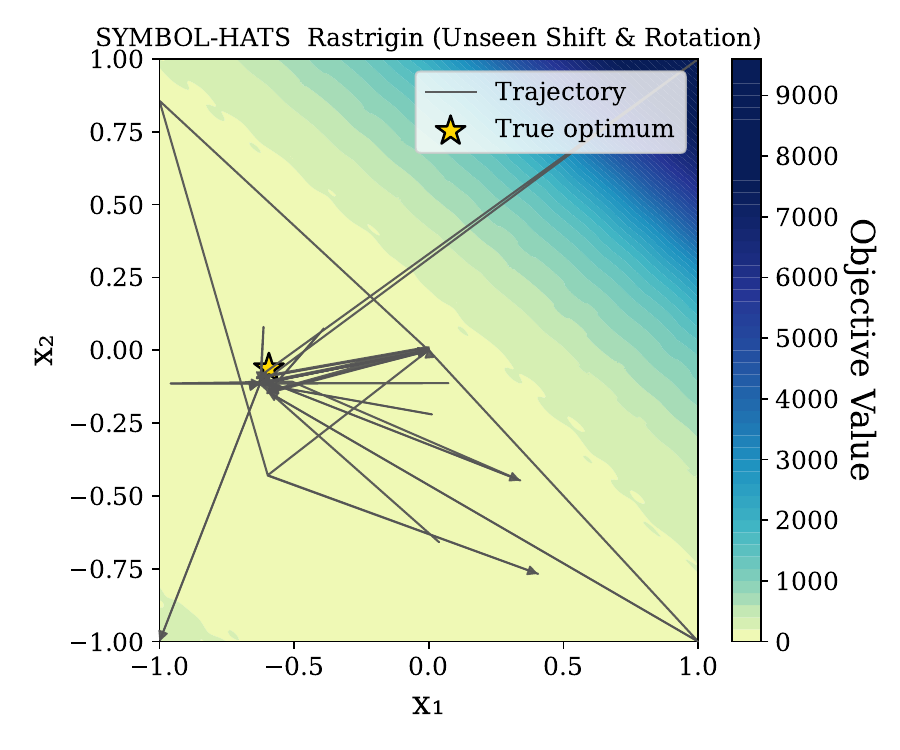}
    \includegraphics[width=0.45\linewidth]{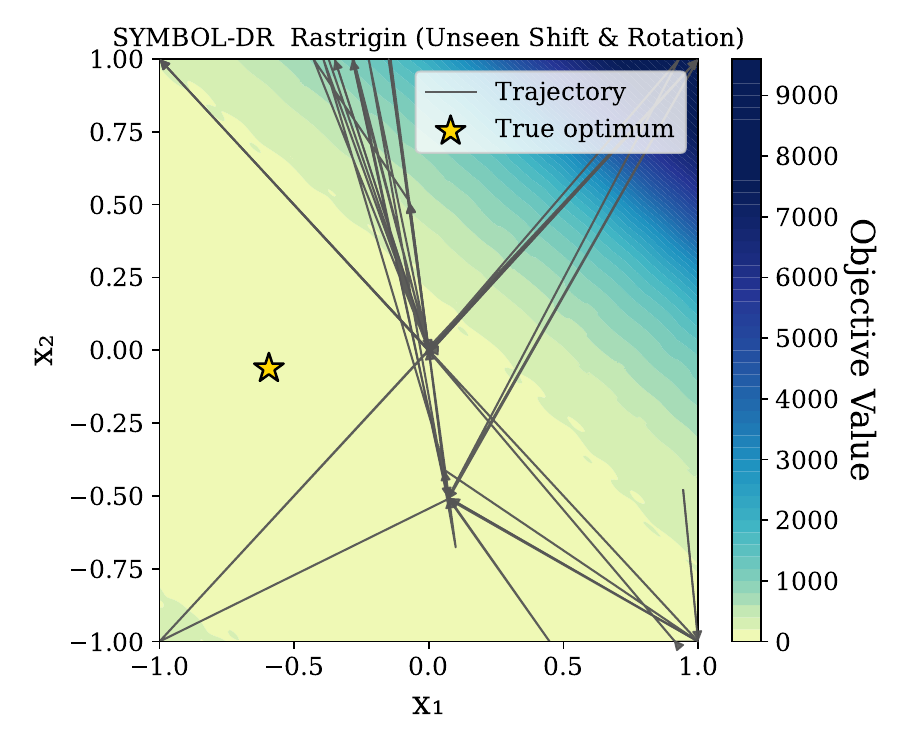} \\
    \includegraphics[width=0.45\linewidth]{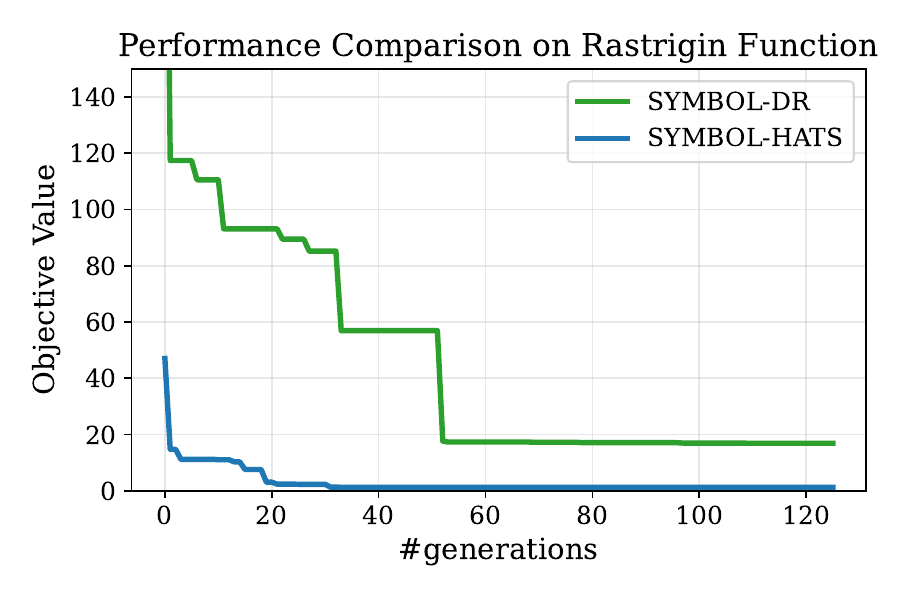}
    
    \caption{Performance comparison on Rastrigin: SYMBOL trained with HATS (termed as SYMBOL-HATS) versus SYMBOL-DR. The top panels visualize the 2D objective landscape of an unseen shifted-and-rotated Rastrigin function with optimization trajectories produced by SYMBOL-HATS (left) and SYMBOL-DR (right), and the bottom panel compares their objective values across generations. SYMBOL-HATS exhibits more stable search behavior.}
    \label{fig:PeformanceComparison}
\end{figure}

 Figure~\ref{fig:PeformanceComparison} compares the behaviors of SYMBOL-DR and SYMBOL-HATS on an unseen shifted-and-rotated 2D Rastrigin instance. In the left panel, SYMBOL-HATS reaches the true optimum within fewer iterations, and its trajectory is more clustered with fewer large cross-region jumps, indicating a more stable search behavior. In contrast, in the right panel, SYMBOL-DR repeatedly searches around $(0,0)$ and makes frequent long-range jumps across the domain, failing to reach the true optimum. Consistent with these observations, the bottom curve shows that SYMBOL-HATS reduces the objective faster, whereas SYMBOL-DR decreases more slowly and levels off at a worse objective.

\subsection{Evaluation under Matched Function-Evaluation Budgets}
\label{appendix:matched-budget}

\begin{figure}[!h]
    \centering
    \includegraphics[width=0.48\linewidth]{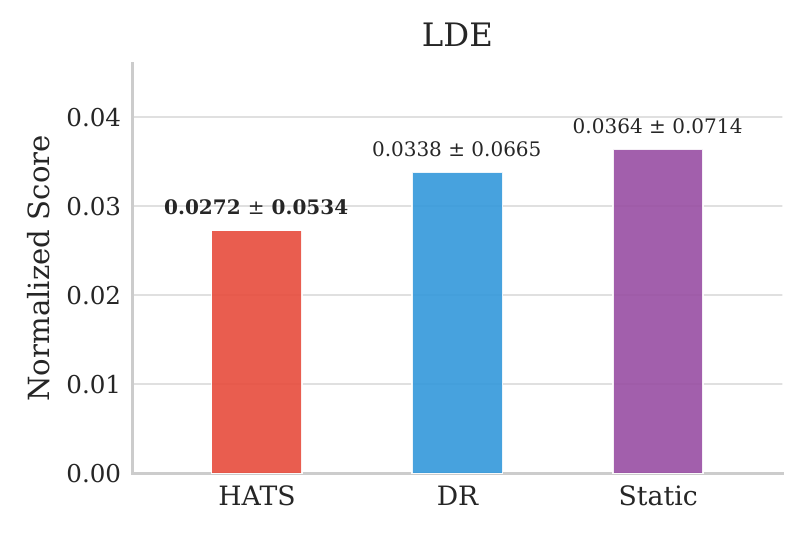}\hfill
    \includegraphics[width=0.48\linewidth]{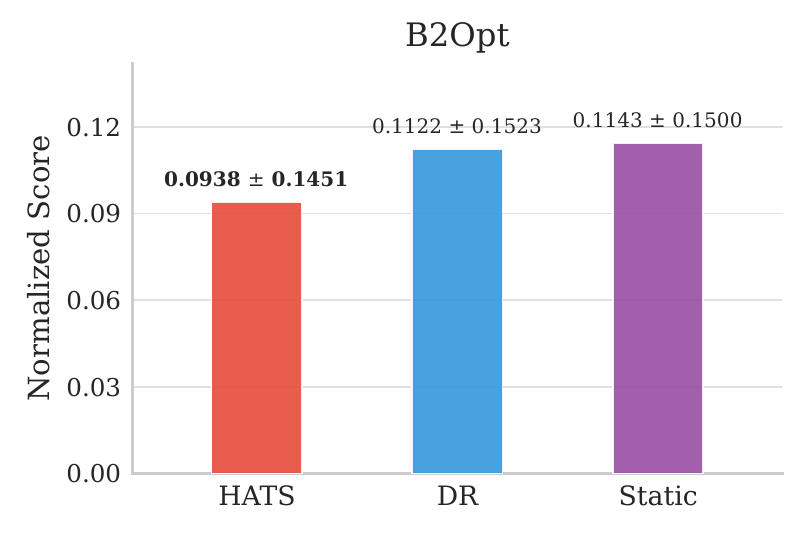}
    \caption{Test performance comparison of HATS, DR, and Static training on CEC using LDE (left) and B2Opt (right) under matched training evaluation budgets (lower is better).}
    \label{fig:matched-budget}
\end{figure}

We further compare different training strategies under the same training-time fitness evaluation (FE) budgets, accounting for both learner rollouts and reference evaluations. The evaluation overhead differs across MetaBBO backbones:
\begin{itemize}[leftmargin=1em, labelindent=0em]
    \item \textbf{Teacher-guided backbones (e.g., SYMBOL):} SYMBOL~\citep{chen2024symbol} naturally runs a teacher optimizer (MadDE) during training to guide its policy. In HATS, we directly reuse these MadDE trajectories already required by SYMBOL's training objective as the reference in Equation~\eqref{eq:gap}, introducing no extra evaluation budget beyond the original method.
    \item \textbf{Teacher-free backbones (e.g., LDE and B2Opt):} Backbones such as LDE~\citep{lde} and B2Opt~\citep{b2opt} do not use a teacher during training, so computing Equation~\eqref{eq:gap} introduces additional reference evaluations. To ensure a fair comparison, we align the total training FEs across methods: we train the two baselines (DR and Static) to their final checkpoints, compute their total training FEs, and select the corresponding HATS checkpoint saved at the matched total training FE budget (summing both learner and reference evaluations) for testing on unseen CEC instances.
\end{itemize}
As shown in Figure~\ref{fig:matched-budget}, under matched training evaluation budgets on CEC, HATS consistently achieves better performance than both DR and Static training on LDE and B2Opt, confirming the effectiveness of HATS even under the same total evaluations on training instances.

\clearpage
\subsection{Interpreting the Prioritized Replay Results in Table~\ref{tab:main-results}}
\label{appendix:hierarchy-scoring}

To better understand the behavior of the prioritized replay baselines in Table~\ref{tab:main-results}, we inspect their training logs and buffer snapshots over the final $40$ epochs (epochs $60$--$99$), measuring effective class coverage via entropy, $N_{\mathrm{eff}} = \exp\bigl(-\sum_{k=1}^{K} p_k \log p_k\bigr)$:

\begin{table}[!h]
\centering
\caption{Class-level training exposure over the final $40$ epochs (epochs $60$--$99$, parameter-update batches) on CEC ($K=10$) and BBOB ($K=24$).}
\begin{tabular}{ll|cc}
\toprule
Base Functions & Training Strategy & Max Single-Class Share & Effective Classes ($N_{\mathrm{eff}}$) \\ \midrule
\multirow{3}{*}{CEC ($K=10$)} & Flat replay & $48.8\%$ (\texttt{Grie\_rosen}) & $5.41$ \\
 & PLR$^{\perp}$-L1 & $26.6\%$ (\texttt{Composition2}) & $7.83$ \\
 & HATS-L1 & $11.4\%$ (\texttt{bi\_Rastrigin}) & $9.98$ \\ \midrule
\multirow{3}{*}{BBOB ($K=24$)} & Flat replay & $31.5\%$ (\texttt{F9}) & $9.92$ \\
 & PLR$^{\perp}$-L1 & $12.5\%$ (\texttt{F7}) & $20.56$ \\
 & HATS-L1 & $4.6\%$ (\texttt{F19}) & $23.98$ \\
\bottomrule
\end{tabular}
\label{tab:hierarchy-exposure}
\end{table}

\begin{figure}[!h]
    \centering
    \includegraphics[width=\linewidth]{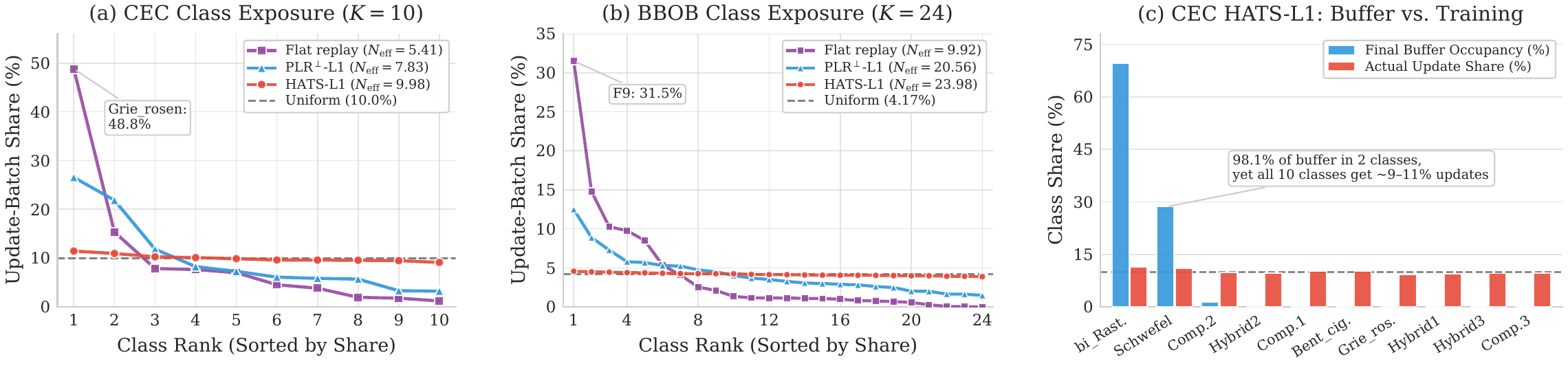}
    \caption{Class-level training exposure over the final $40$ epochs: (a--b) sorted class update shares on CEC and BBOB; (c) final buffer occupancy vs.\ actual update shares across CEC classes under HATS-L1.}
    \label{fig:coverage-decoupling}
\end{figure}

\begin{figure}[!h]
    \centering
    \includegraphics[width=0.92\linewidth]{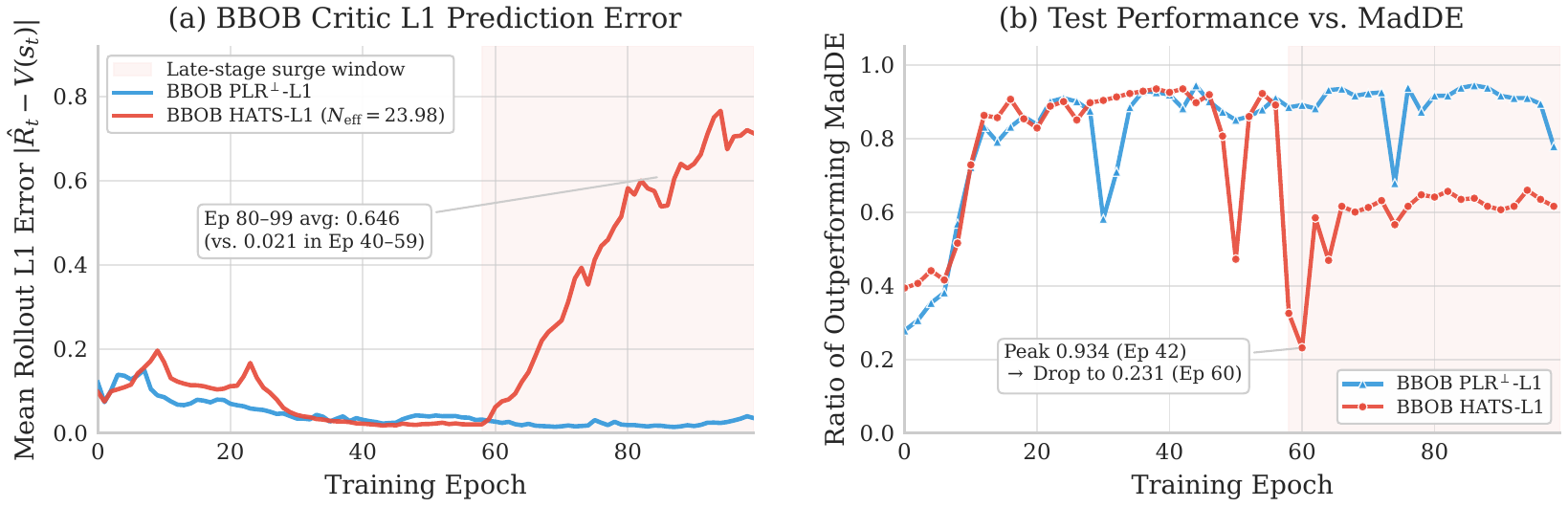}
    \caption{Training dynamics of PLR$^{\perp}$-L1 and HATS-L1 on BBOB: (a) mean critic L1 error $|\hat{R}_t - V(s_t)|$, and (b) test performance vs.\ MadDE.}
    \label{fig:bbob-l1-dynamics}
\end{figure}

\begin{itemize}[leftmargin=1em, labelindent=0em]
    \item \textbf{Hierarchy protects actual training coverage rather than buffer uniformity.} As shown in Figure~\ref{fig:coverage-decoupling}(a--b) and Table~\ref{tab:hierarchy-exposure}, removing class scheduling causes Flat replay to skew updates heavily toward a few dominant classes on both CEC and BBOB, whereas hierarchical scheduling maintains broad coverage across classes. Notably, in CEC HATS-L1 (Figure~\ref{fig:coverage-decoupling}(c)), even though \texttt{bi\_Rastrigin} and \texttt{Schwefel} dominate almost the entire final buffer, actual training updates remain balanced across all $10$ classes---because HATS selects the class first and generates a new instance whenever that class has no cached entry.
    \item \textbf{BBOB HATS-L1 degrades due to late-stage instability under value-loss scoring, not lack of diversity.} Although BBOB HATS-L1 covers all $24$ classes almost uniformly ($N_{\mathrm{eff}}=23.98$), its mean L1 error surges from $0.0215$ (epochs $40$--$59$) to $0.6458$ (epochs $80$--$99$), whereas PLR$^{\perp}$-L1 stays stable; meanwhile, its ratio of outperforming MadDE drops significantly, as shown in Figure~\ref{fig:bbob-l1-dynamics}. Prior UED studies also show that critic value-loss priorities are prone to non-stationarity, stochasticity, and misalignment with true learnability~\citep{robust-plr,samplr,no-regrets-ued}---a class harder for the critic to fit is not necessarily more valuable to train on, especially across $24$ heterogeneous BBOB classes. In contrast, our reference--agent gap anchors each task against a fixed reference optimizer (MadDE), offering a more stable cross-class signal.
\end{itemize}

\clearpage

\subsection{Optimization Curve}
\label{appendix:curve}

\begin{figure}[!h]
    \centering
    \includegraphics[width=\linewidth]{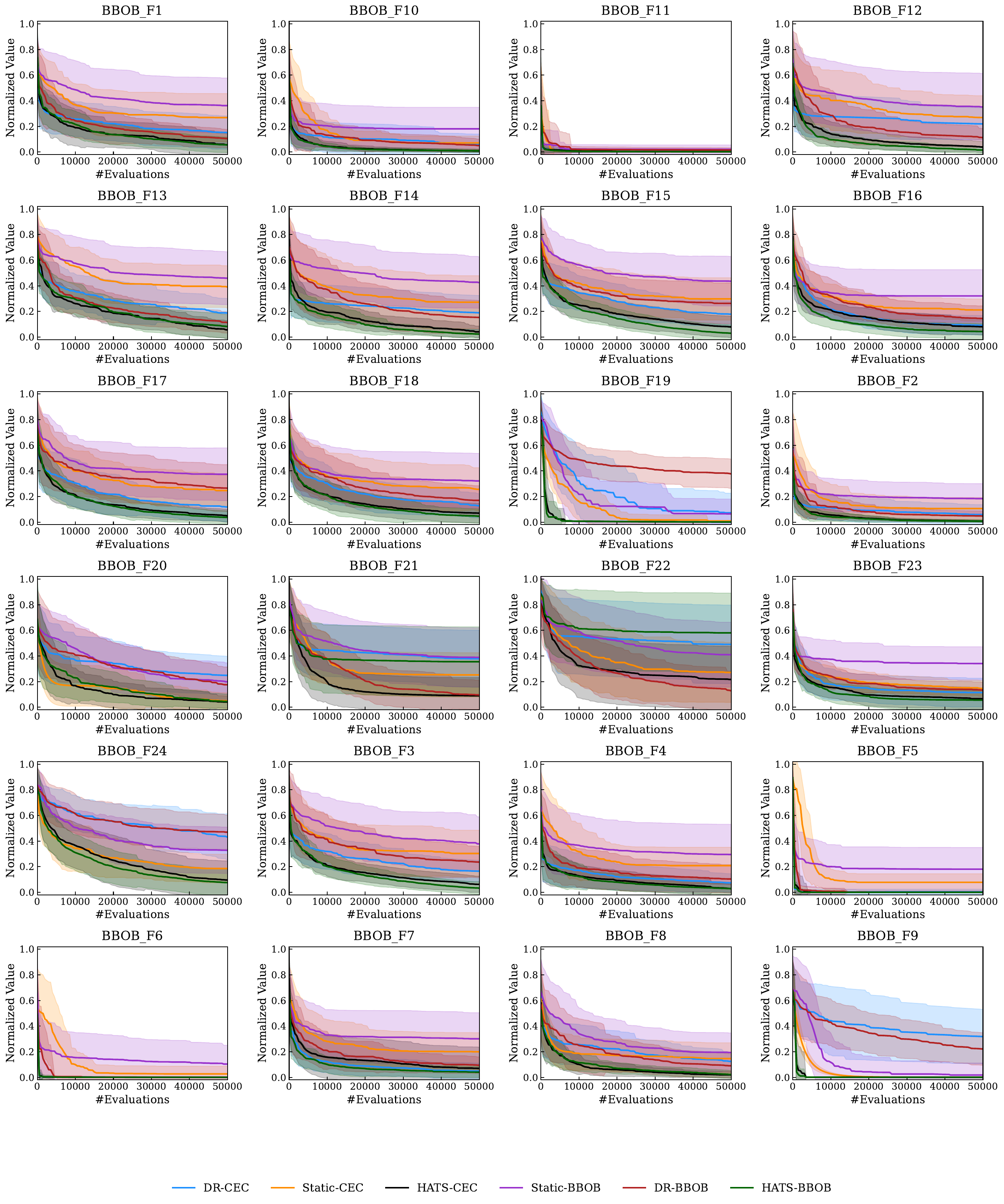}

    \caption{Optimization curve of different methods on BBOB.}
    \label{fig:bbob-f3}
\end{figure}

\begin{figure}[!h]
    \centering
    \includegraphics[width=\linewidth]{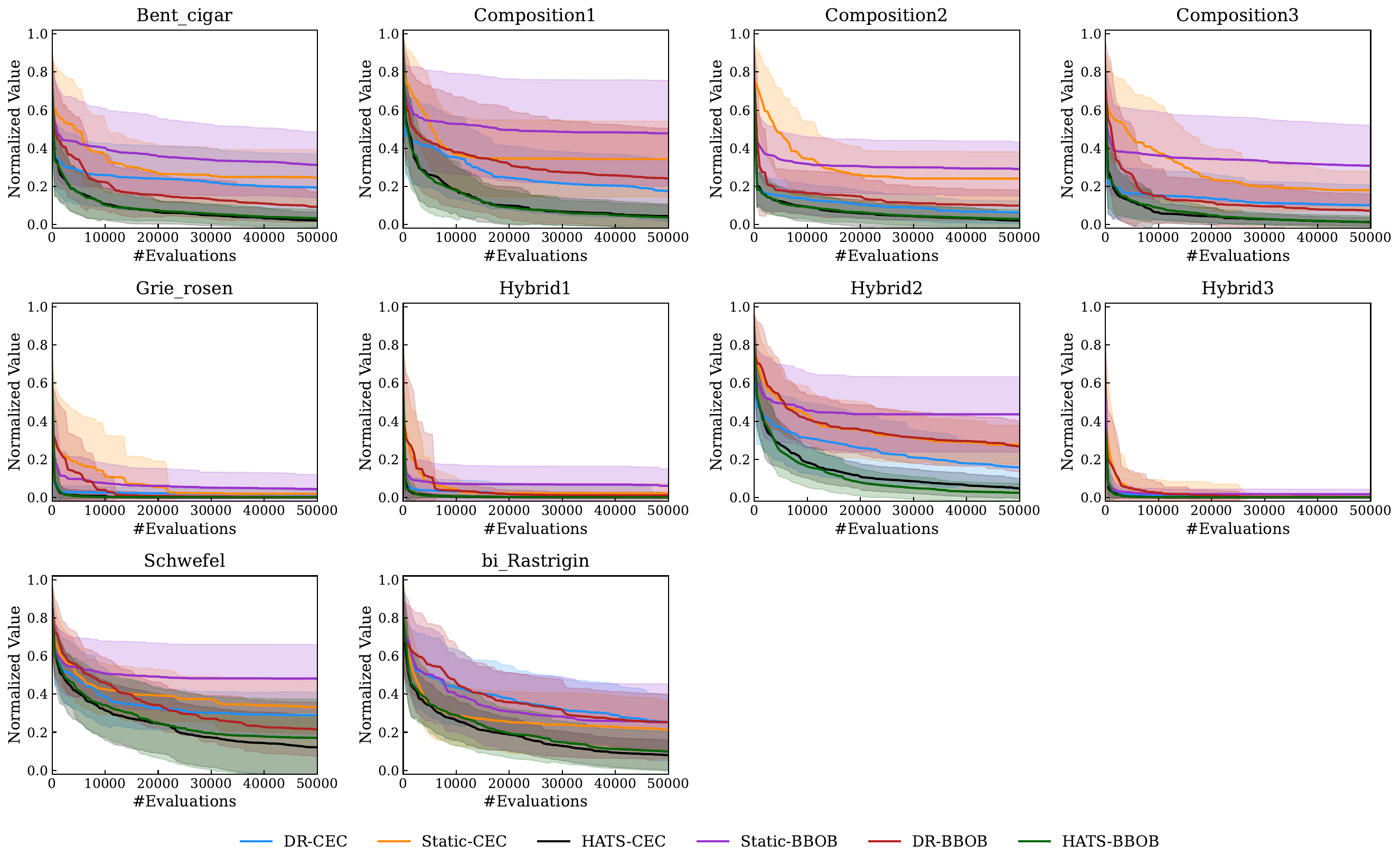}
    \caption{Optimization curve of different methods on CEC.}
    \label{fig:cec}
\end{figure}

\begin{figure}[!h]
    \centering
    \includegraphics[width=0.49\linewidth]{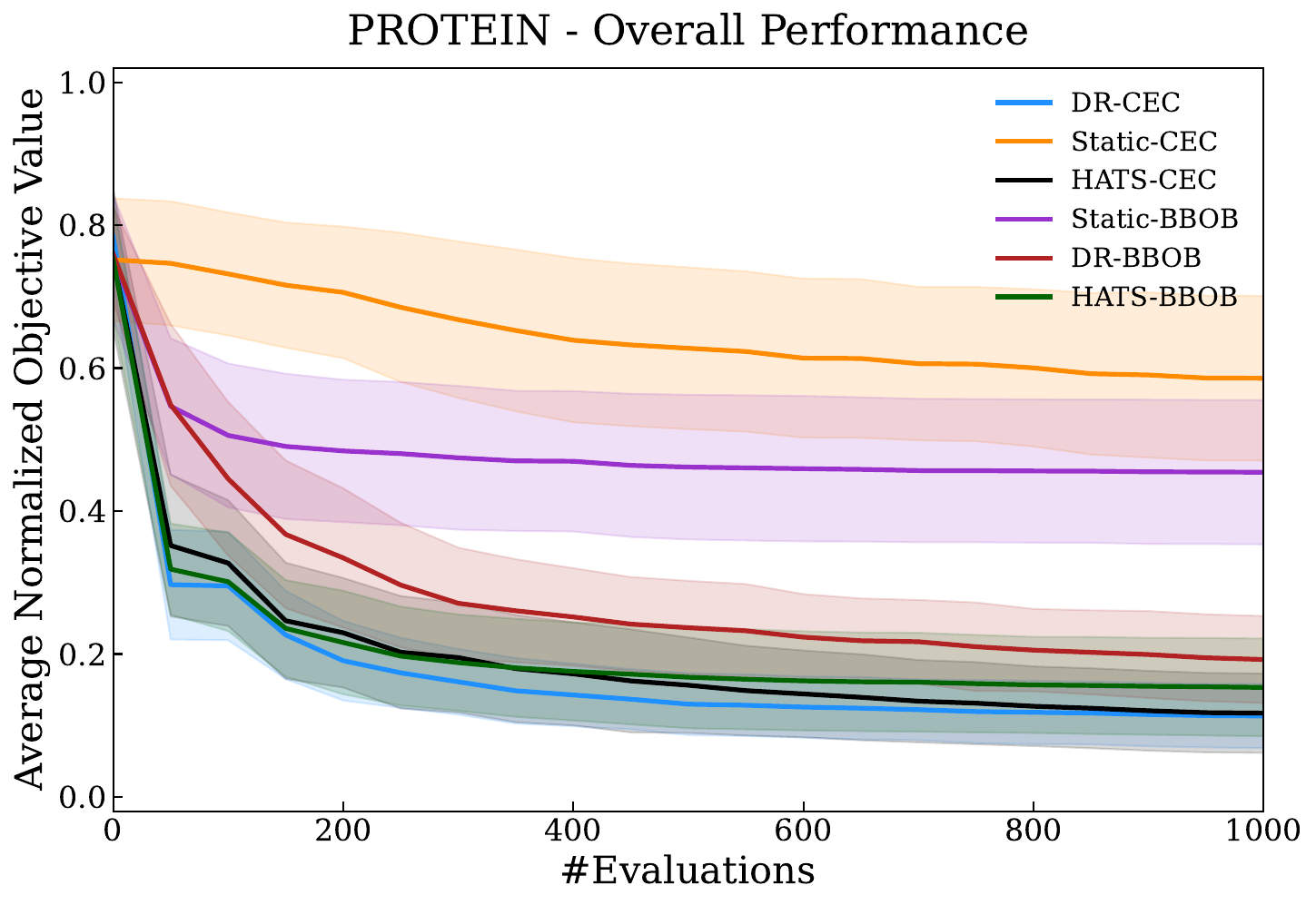}
    \includegraphics[width=0.49\linewidth]{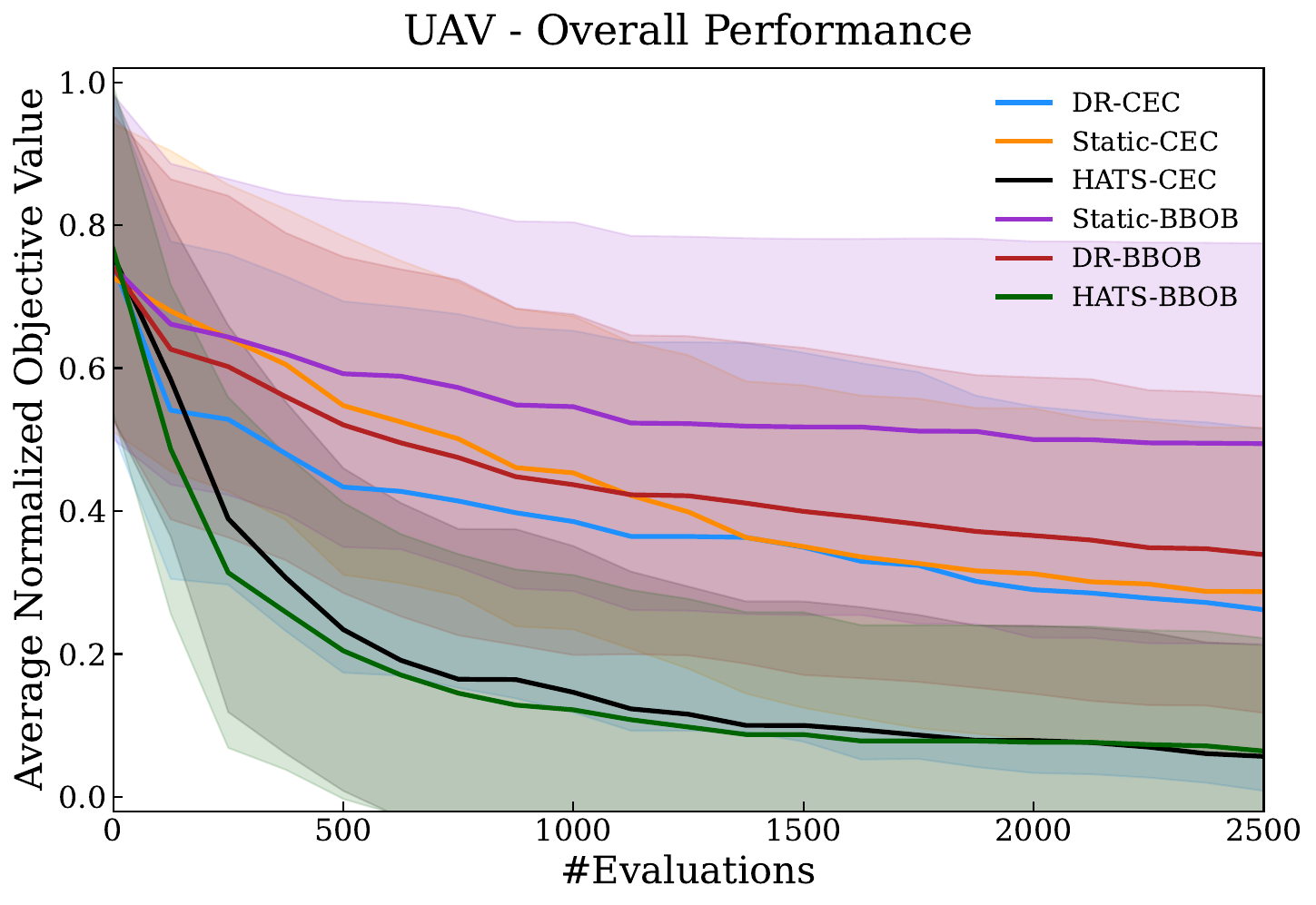}
    \caption{Optimization curve of different methods on Protein Docking and UAV.}
    \label{fig:uav}
\end{figure}

In~\cref{fig:bbob-f3,fig:cec,fig:uav}, we visualize the optimization curve of all compared methods on BBOB, CEC, Protein and UAV.
Note that for BBOB and CEC, we augment the testing instances by randomly sampling unseen parameters for testing, we plot the detailed performance of each function by averaging objective value at each iteration.
For protein docking and UAV, we visualize the average performance over all test problems.
The objective value are normalized to [0,1] with respect to each instance.
These results clearly show the superiority of our proposed method, HATS.

\end{document}